\documentclass[pdflatex,sn-mathphys-num]{sn-jnl}

\usepackage{graphicx}
\usepackage{multirow}
\usepackage{amsmath,amssymb,amsfonts}
\usepackage{amsthm}
\usepackage{mathrsfs}
\usepackage[title]{appendix}
\usepackage{xcolor}
\usepackage{textcomp}
\usepackage{manyfoot}
\usepackage{booktabs}
\usepackage{algorithm}
\usepackage{algorithmicx}
\usepackage{algpseudocode}
\usepackage{listings}

\usepackage[T1]{fontenc}
\usepackage{caption}
\DeclareCaptionFormat{ext}{\textbf{Extended Data #1}#2#3}

\usepackage[normalem]{ulem}
\usepackage{siunitx}    
\usepackage{comment}
\usepackage{adjustbox}
\usepackage{tabularx,array}

\usepackage{svg}

\providecommand{\pinav}[1]{\pi_{\text{nav}}(#1)}
\providecommand{\piloc}[1]{\pi_{\text{loc}}(#1)}

\theoremstyle{thmstyleone}

\theoremstyle{thmstyletwo}

\theoremstyle{thmstylethree}

\begin{document}

\title{Asymmetric physics enables efficient learning in quadrupedal robot swarms}

\author[1]{\fnm{Yuang} \sur{Zhang}}
\equalcont{These authors contributed equally to this work.}

\author[2]{\fnm{Yunlong} \sur{Song}}
\equalcont{These authors contributed equally to this work.}

\author[1]{\fnm{Zhihao} \sur{He}}
\equalcont{These authors contributed equally to this work.}

\author[1]{\fnm{Zelin} \sur{Ni}}
\equalcont{These authors contributed equally to this work.}

\author[1]{\fnm{Kangyu} \sur{Wang}}

\author[1]{\fnm{Tianchi} \sur{Liu}}

\author[1]{\fnm{Yu} \sur{Hu}}

\author[1]{\fnm{Feng} \sur{Yu}}

\author*[1]{\fnm{Danping} \sur{Zou}}
\email{dpzou@sjtu.edu.cn}

\author*[1]{\fnm{Weiyao} \sur{Lin}}
\email{wylin@sjtu.edu.cn}

\affil[1]{\orgname{Shanghai Jiao Tong University}, \orgaddress{\city{Shanghai}, \country{China}}}

\affil[2]{\orgname{Independent Researcher}}

\abstract{

Animal collectives navigate cluttered environments through local coordination, yet robot swarms still struggle to reproduce this capability in the physical world. End-to-end learning offers a route to such coordination, but scaling it to embodied swarms remains difficult: standard sampling-based reinforcement learning becomes inefficient when visual perception, dense robot–robot interaction, and contact-rich locomotion must be learned together. Here we show that asymmetric physics enables efficient end-to-end learning of vision-based, decentralized control in large swarms of quadrupedal robots. During training, quadrupeds interact in shared environments, where a high-fidelity, non-differentiable simulator generates realistic motion and contact dynamics, and differentiable surrogate models provide gradients for navigation and locomotion policies. This separation enables up to 512 quadrupeds to learn coordinated navigation policies in obstacle-rich environments. At deployment, each robot acts from a single forward-facing depth camera, without explicit communication, centralized planning, or global maps. The policies generalize across forests, bridges, enclosures, narrow passages, and mazes, and zero-shot transfer to six physical quadrupeds across five real-world scenarios. The resulting swarms exhibit predictive avoidance, right-side yielding, pausing before bottlenecks, and wall following, showing that asymmetric physics enables efficient training of scalable decentralized control policies for quadrupedal robot swarms.
}


\maketitle

\subsection*{INTRODUCTION}
A herd of sheep can move through a dense forest, cross a narrow bridge, and reach an open pasture while avoiding obstacles, maintaining spacing, yielding to one another, and adapting its motion to the terrain. No single animal controls the group. Instead, coordinated motion emerges from local perception and interaction. Such self-organized coordination has long inspired swarm robotics~\cite{kube1993collective,beni2004swarm,csahin2004swarm}. If reproduced in robots, it could enable large collectives to explore planets~\cite{kang2019marsbee,arm2023scientific}, monitor environments~\cite{bayat2017environmental,dunbabin2012robots}, search for targets~\cite{horyna2023decentralized}, or operate in disaster sites where communication is unreliable and individual failures are expected. Despite decades of progress in robotics and artificial intelligence, building robot swarms that approach the adaptability of animal collectives remains a major challenge, especially in real-world settings~\cite{yang2018grand,dorigo2020reflections}. Contact-rich mobile robots provide a particularly demanding test case for this challenge. Among them, quadrupedal robots are demanding because their motion depends on frequent ground contact, high-dimensional whole-body dynamics, and tight coupling between perception, navigation, and locomotion. A training principle that succeeds in this setting may therefore inform the design of broader classes of embodied robot swarms, including other legged and mobile platforms.

Classical swarm models and early robot collectives have shown that local rules can generate aggregation, flocking and other collective behaviors~\cite{reynolds1987flocks,vicsek1995novel,couzin2005effective,mondada2009puck,rubenstein2014programmable}. However, these systems typically rely on simplified sensing, planar motion, global state information, or reliable communication. Such assumptions become restrictive for contact-rich mobile swarms in cluttered environments, where each robot must infer nearby agents and obstacles from onboard sensors while maintaining stable contact-rich locomotion.

Robot learning offers a route beyond hand-designed swarm rules.
Imitation learning (IL)~\cite{osa2018algorithmic} and reinforcement learning (RL)~\cite{sutton2018reinforcement} are widely used to train robotic control policies.
IL requires expert demonstrations, which are difficult to obtain for large collectives, and policies distilled from hand-designed swarm controllers inherit the limitations of those controllers~\cite{kuckling2023recent}.
RL has achieved strong results in single-robot control, including legged locomotion~\cite{hwangbo2019learning,miki2022learning}, autonomous racing~\cite{fuchs2021super,wurman2022outracing}, and high-speed drone flight~\cite{kaufmann2023champion,song2023reaching}.

Although multi-agent RL provides a natural framework for decentralized coordination~\cite{huh_multi-agent_2023,orr_multi-agent_2023,chung_learning_2024,wang_distributed_2022}, few methods have been successfully applied to real-world robot swarms.
A central obstacle is training efficiency: the joint state-action space grows with the number of robots, and visual observations introduce high-dimensional inputs for every agent.
For contact-rich platforms, this difficulty is compounded by discontinuous contacts, high-dimensional body dynamics, and low-level control. Legged swarms represent an especially demanding instance, because each robot must maintain stable locomotion while negotiating dense interactions with other robots and obstacles. As a result, model-free, sampling-based RL is already costly for a single vision-based robot and becomes inefficient when dozens or hundreds of contact-rich robots must learn through shared interaction.
Moreover, swarm control cannot be obtained by simply copying a single-robot policy to many robots. A single-robot visual policy may learn obstacle avoidance, but it does not experience the dense robot-robot interactions required for yielding, pausing before conflict passages, passing through bottlenecks, and resolving congestion. Learning such behaviors requires efficient training in shared multi-robot environments, where coordination emerges from repeated local interactions rather than hand-designed rules or centralized supervision.

These observations point to a broader training principle: large-scale interaction should not be treated only as stochastic samples for trial-and-error optimization, but also as a physical structure from which useful first-order optimization signals can be extracted. Biological systems suggest that physical reasoning can reduce exploration and support learning from limited experience~\cite{piaget1954construction,baillargeon2008innate,garrido2025intuitive,ullman2017mind}.
In robotics, first-principles physics can provide first-order gradients that often have lower variance than zeroth-order estimates used in model-free RL, improving stability and convergence speed~\cite{suh2022differentiable,xu2021accelerated}. However, directly differentiating through realistic contact, collision, and locomotion remains difficult for complex mobile robots, and applying physics-based gradient methods to large robot swarms remains largely unexplored. This creates a tension: realistic swarm training requires high-fidelity simulation of contact-rich multi-robot dynamics, whereas efficient learning requires smooth optimization signals.

Here, we instantiate this principle in quadrupedal robot swarms, a demanding class of contact-rich embodied systems, with an efficient end-to-end training framework for vision-based swarm control. We study a navigation task in which each robot must reach its own goal using onboard vision while avoiding static obstacles and other robots. Each robot observes the world through a forward-facing depth camera and knows its own goal and internal state, but does not rely on explicit state estimates of other robots, global maps, centralized planning, or inter-robot communication. This setting captures a core challenge for embodied swarm autonomy: local visual perception, multi-robot interaction, and contact-rich legged locomotion are tightly coupled.

The key idea is not to make the entire robot-swarm simulator differentiable, but to separate the physics used for behavior generation from the physics used for learning. During forward rollouts, a high-fidelity non-differentiable simulator models realistic quadruped motion, contacts, collisions, and robot-environment interactions. During backpropagation, simple differentiable surrogate models provide smooth gradients for policy optimization, allowing competing objectives such as speed tracking, collision avoidance, and passage through narrow spaces to shape individual actions more directly. Specifically, point-mass and rigid-body surrogate models approximate multi-robot interaction and quadruped locomotion, respectively. The surrogate models are therefore platform-specific approximations of a more general principle: use the simplest differentiable physics that captures the dominant interaction and locomotion structure, while retaining high-fidelity simulation for behavior generation. These approximations are used only for gradient computation, whereas the forward rollouts remain grounded in high-fidelity simulation. This separation preserves contact-rich realism while avoiding unstable gradients from discontinuous dynamics.

The framework trains a hierarchical policy stack that links vision, navigation, and locomotion. A high-level navigation policy maps depth observations and goal information to velocity commands, and a low-level locomotion policy converts these commands into joint-level actions. This design allows visual swarm navigation and legged locomotion control to be trained within one framework. Crucially, it allows hundreds of quadrupeds to interact during training in the same shared environment, rather than learning only from isolated single-robot experience. \emph{This dense interaction is essential for learning swarm behaviors: in our experiments, single-agent visual policies transfer poorly to crowded multi-robot scenes, producing less ordered passage, less consistent spacing, more turnover or timeout events, and lower completion rates as swarm size increases.}

Using this framework, we train decentralized vision-based policies for large quadruped swarms as a challenging demonstration of the principle. In simulation, a single learned policy controls up to $512$ quadrupeds in shared, obstacle-rich environments, including dense forests, narrow groves, opposing bridge crossings, fenced enclosures, circular exits, and intersecting-path mazes.
The learned policy transfers directly to six physical Unitree Go2 quadrupeds. We deploy it without additional fine-tuning, with each robot equipped with a forward-facing depth camera. Across five real-world scenarios, including a forest, a narrow bridge, a single-robot-width gap, a cluttered pavilion, and an obstacle-dense room, each robot acts independently using local visual input and its own goal. Despite the absence of explicit communication, centralized coordination, global maps, or hand-designed group-level rules, the robots demonstrate learned multi-robot navigation behaviors. These include predictive avoidance of nearby agents, right-side yielding in head-on encounters, pausing in open areas before conflict passages, and wall-following under limited perception. These behaviors are neither pre-programmed nor induced by group-level objectives, but emerge from local sensing, physical interaction, and individual goal-directed learning.

The real-world and simulation results provide, to our knowledge, the first demonstration of vision-based learned swarm navigation in quadrupedal robots under decentralized sensing, contact-rich locomotion, and dense multi-robot interaction. The novelty lies not in the individual behaviors themselves, but in how they are learned: through efficient end-to-end visual training in shared multi-robot environments, without communication, centralized planning, or hand-designed coordination rules. More broadly, they suggest a training principle for contact-rich embodied swarm control: when realistic contact dynamics are difficult to differentiate directly, separating high-fidelity forward simulation from differentiable surrogate optimization can turn large-scale interaction from a sampling bottleneck into an efficient training signal. Quadrupedal robots provide a demanding test case for this principle. The same separation between realistic forward physics and differentiable surrogate physics may inform training methods for other contact-rich platforms, including tracked robots and other legged machines.

\subsection*{Related work}

Classical models such as Reynolds’ Boids model~\cite{reynolds1987flocks}, the Vicsek model~\cite{vicsek1995novel}, and biologically inspired interaction rules~\cite{couzin2005effective} show that simple local rules can generate emergent group-level coordination. Nevertheless, these models abstract away the sensing, dynamics, contacts, and environmental variability faced by real robots. Pipelines based on hand-designed perception-planning-control modules and explicit inter-robot communication can be effective~\cite{vasarhelyi2018optimized, zhou2022swarm}, but they remain difficult to scale because interaction rules must be tuned across environments, and communication requirements grow rapidly with swarm size.

Early research explored automatic design of swarm controllers using evolutionary algorithms to tune neural networks~\cite{trianni2008evolutionary} or modular structures such as finite-state machines~\cite{francesca2014automode} and behavior trees~\cite{kuckling2022automode}. Robot collectives such as e-pucks~\cite{mondada2009puck}, Kilobots~\cite{rubenstein2014programmable}, and Khepera robots~\cite{mondada1999development} demonstrated clustering, aggregation, flocking, and other collective behaviors in controlled settings~\cite{gauci2014self, garnier2005aggregation}. There are also impressive real-world deployments of robot swarms~\cite{duarte2016evolution, vallegra2018gradual}. However, many of these systems assume global state information or rely on simplified sensing and locomotion models (e.g., precise position measurements, 2D range-and-bearing sensing, planar kinematics, or reliable communication links such as WiFi). These assumptions become restrictive for legged robot swarms in cluttered environments, where robots must infer nearby agents and obstacles from onboard sensors while maintaining stable contact-rich locomotion.

Deep reinforcement learning (RL) has been widely explored for multi-agent control tasks, demonstrating promising results in both coordination and emergent behaviors.
Notable studies include the hide-and-seek game~\cite{baker2019emergent} using PPO, defeating world champions in the game of Dota 2 using large-scale RL~\cite{berner2019dota}, and team play in simulated humanoid football~\cite{liu2022motor}.
Deep reinforcement learning (RL) has been widely explored for multi-agent coordination and has produced complex emergent behaviors in simulated domains~\cite{long2017deep, sartoretti_primal_2019, bansal_emergent_2018}.
However, most learned policies remain validated in constrained settings, either simulated or on overly simplified platforms, with limited attention to realistic sensing, complex dynamics, and physical interactions.
Existing multi-agent RL methods have therefore seen limited deployment in real-world robot swarms, especially at large scale and with contact-rich locomotion~\cite{huh_multi-agent_2023, chung_learning_2024}.

A central difficulty is training efficiency. As the number of robots increases, the joint state-action space grows rapidly, while high-dimensional sensory inputs, such as depth images, increase the difficulty of policy optimization.
This challenge is further exacerbated in scenarios involving high-dimensional sensory inputs (e.g., depth maps), discontinuous contacts, whole-body dynamics, and low-level locomotion control.
To address these limitations, recent research has increasingly focused on incorporating model priors, either through first-principles physics-based simulation frameworks~\cite{freeman2021brax, howelllecleach2022, ren2022diffmimic} or via learned world models~\cite{hafner2025mastering, wu2023daydreamer, hansen2022temporal, li2025robotic}.
In particular, differentiable physics simulation (DPS) offers a promising avenue by embedding robot dynamics directly into the learning loop, enabling the computation of first-order policy gradients through backpropagation.
While DPS yields low-variance gradients and often accelerates convergence compared to traditional RL, its application to contact-rich systems remains fraught with difficulties, including gradient discontinuities and optimization instability~\cite{song_learning_2024, wiedemann2023training}.
Notably, recent advances have demonstrated the feasibility of vision-based quadrotor policy learning using DPS~\cite{zhang2025learning}. However, extending these approaches to quadrupeds operating in cluttered and dynamic swarm environments continues to be a critical and largely unresolved challenge.

\begin{figure}[thbp!]
    \centering
    \includegraphics[width=1.0\linewidth]{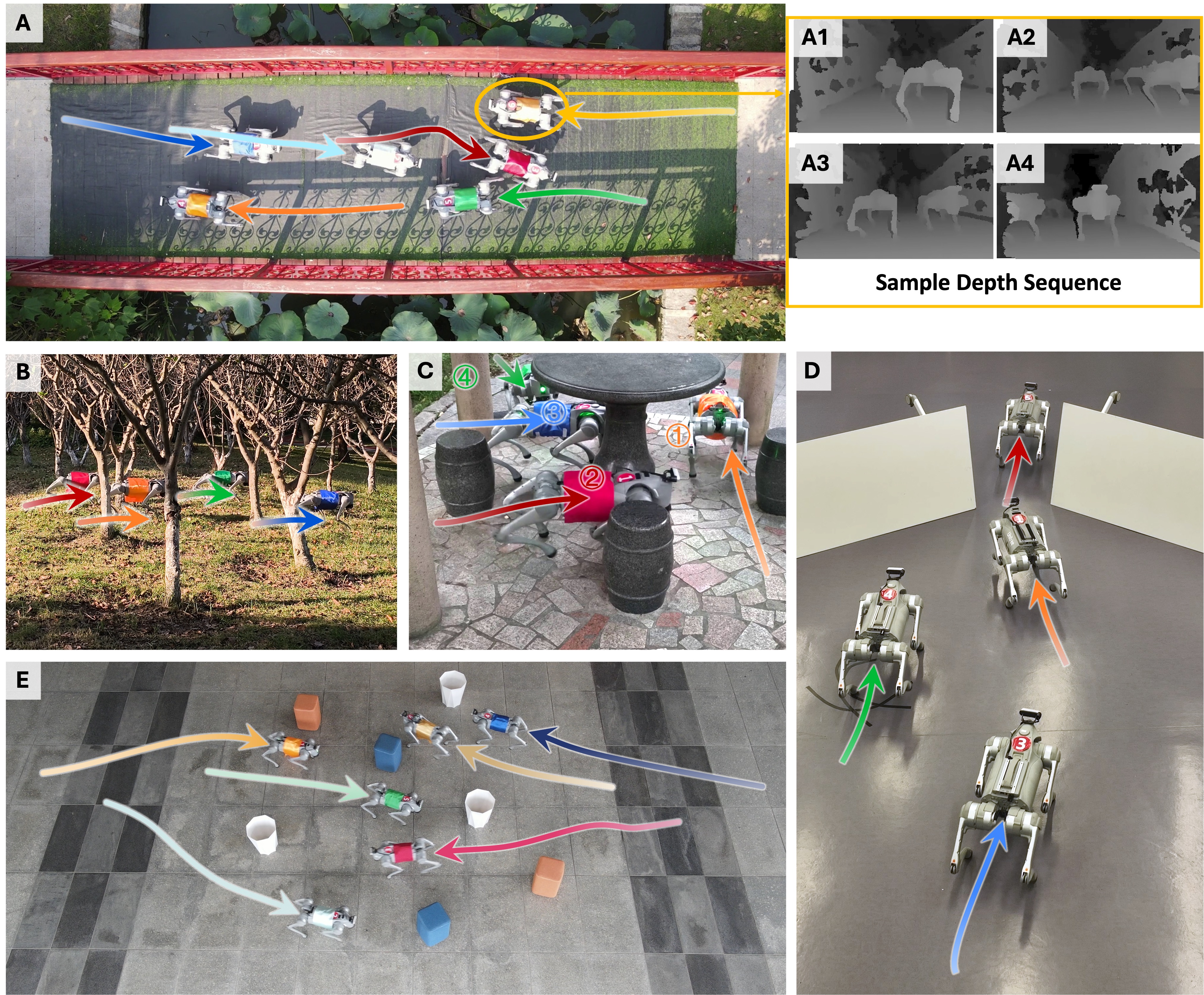}
    \caption{\textbf{Real-world demonstrations of vision-only multi-robot navigation.} Six Unitree Go2 quadrupeds, each using only a forward-facing Intel D435i depth camera and an identical onboard policy (no explicit communication between robots), execute goal-directed navigation in diverse, cluttered environments. Shown scenarios include: (A) two teams meeting on a narrow bridge, (B) traversal through a forest with natural obstacles, (C) passage through a pavilion with confined space and furniture-like obstacles,
    {\color{black}(D) passage through a single-robot-width gap, }
    and (E) navigation in a cluttered room with randomly placed obstacles. Supplementary Video~1 provides full trajectories.
    }

    \label{fig:real}
\end{figure}
\newpage

\section*{RESULTS}
This section presents the experimental results of our end-to-end training framework and vision-based policy. We demonstrate its effectiveness in both real-world and simulated scenarios.
Our evaluation focuses on the system's ability to coordinate large robot swarms across diverse environments, including real-world forests and simulated mazes. We analyze emergent coordination behaviors, assess scalability with increasing swarm sizes, and compare our learning approach to state-of-the-art reinforcement learning (RL) methods.

\subsection*{Swarm control of quadrupedal robots in the real world}
We first demonstrate the effectiveness of our trained policy in diverse real-world scenarios.
Specifically, we deployed six Unitree Go2 quadruped robots, each equipped with a forward-facing Intel D435i depth camera.
We tested our policy across a series of diverse and challenging environments.
Emulating the autonomous coordination of animals in the wild, these robots relied exclusively on visual input that has a limited field of view, processing depth information to make real-time decisions without any communication between robots.
We tested our robot team with five different real-world scenarios, designed to evaluate different aspects of their navigation abilities (Figure~\ref{fig:real}):

In the forest setting (B), four quadruped robots are arranged in a 2$\times$2 formation on one side of a row of trees and moved independently toward their respective target points on the other side. Operating in a sequential formation, the robots dynamically adjusted their speeds and trajectories to maintain safe inter-robot distances while maneuvering around obstacles such as scattered trees, small rocks, and uneven terrains.
This scenario highlighted their obstacle avoidance and adaptive terrain traversal capabilities within a realistic, unstructured outdoor setting.
Furthermore, the low placement of their depth cameras, close to the forest ground, introduced significant disturbances caused by thin grasses and low vegetation, posing additional challenges to the robustness and reliability of our vision-based navigation policies. Nevertheless, the robots demonstrated their resilience to unknown disturbances.

We further tested navigation within highly constrained spaces, including a cluttered pavilion (C) and a narrow gap (D).
In the pavilion scenario (C), four quadruped robots were positioned from three different directions. With a stone table and four small stone stools as obstacles in the center, the available space for passage was quite narrow. The robots needed to navigate through the stone stools to reach the exit on the other side.
Here, robots occasionally paused to allow others to move through, which effectively reduced collisions and enabled smooth passage.
Similarly, in the narrow gap scenario (D), which was only wide enough for a single robot, {\color{black}the four quadruped robots started on one side and needed to reach their respective targets on the other, requiring them to alternate passage by adjusting their speeds or pausing briefly}.
{\color{black}While no explicit coordination rules were programmed, these interactions resulted in alternating passage and safe traversal, underscoring}
the emergent cooperative behaviors enabled by our vision-based navigation policy, showcasing its ability to handle complex spatial constraints.

To evaluate the system's ability to coordinate in scenarios with dynamic constraints, we conducted experiments involving both bridge crossing (A) and obstacle-dense room navigation (E).
In the bridge crossing scenario (A), two teams of three robots approach each other from opposite ends of a narrow bridge, mimicking a single-lane road.
Despite the spatial constraints, the robots successfully coordinated their movements using only visual cues to safely exchange positions.
In the cluttered room with randomly placed obstacles (E), we observed robots adjusting their paths and occasionally pausing, which reduced congestion and allowed both groups to reach their goals.
These experiments highlight that, while each robot acts on its own perception and goal, repeated local interactions give rise to consistent coordination behaviors allowing efficient navigation in challenging real-world environments.

\subsection*{Vision-based mutual avoidance in dense swarms}

\begin{figure}[htbp!]
    \centering
    \includegraphics[width=1.0\linewidth]{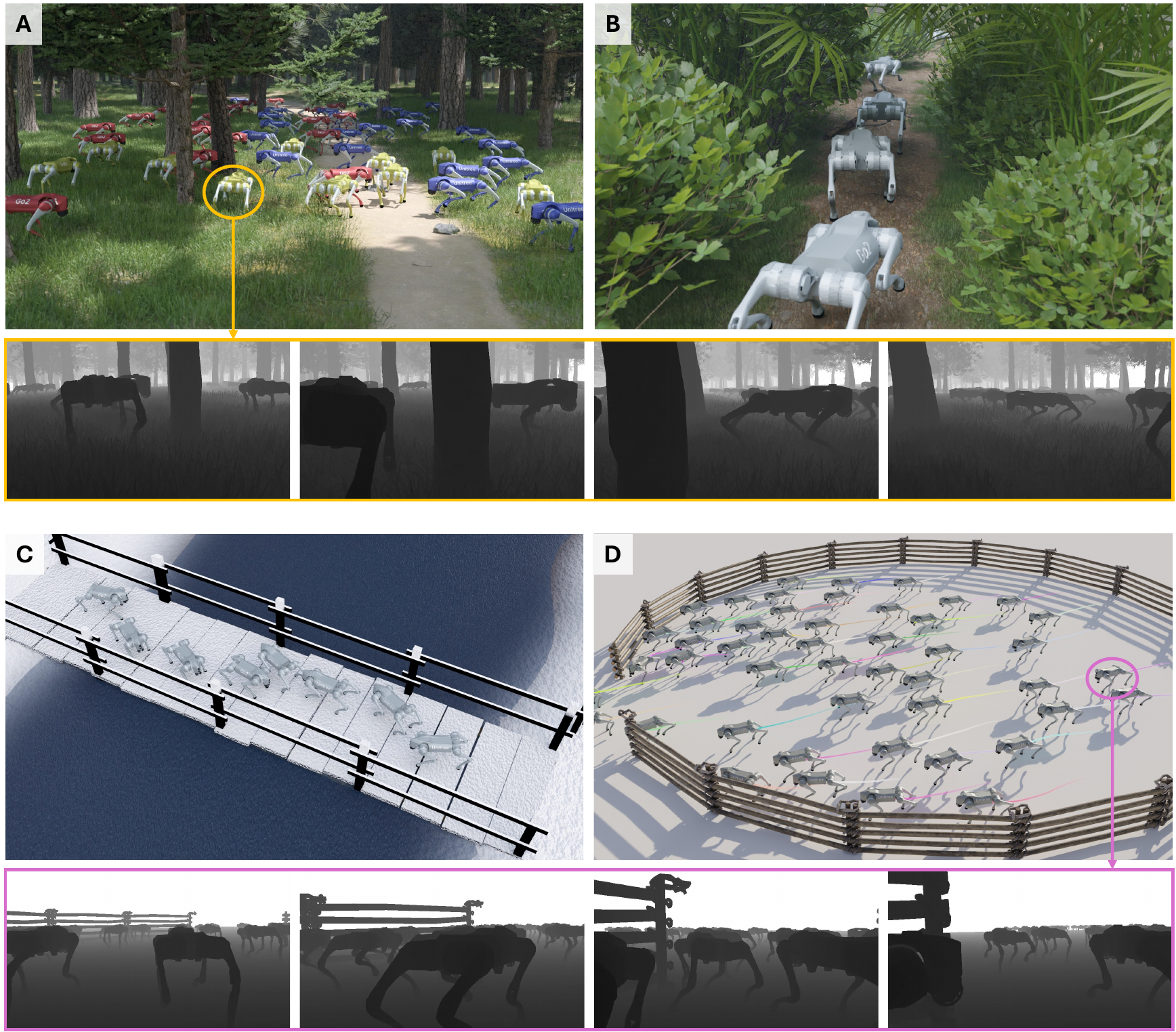}
    \caption{\textbf{Vision-based mutual avoidance in dense swarms.} Qualitative simulation snapshots showing that a single decentralized, vision-only policy scales to dense multi-robot settings and diverse cluttered environments. Scenarios include: (A) Navigating intersecting paths within a dense forest, (B) Forming a line to traverse a narrow grove path, (C) Coordinating a bridge crossing from opposing directions, and (D) Exiting a fenced enclosure through a single narrow opening. All robots operate independently with their own depth camera observation. Videos are available in Supplementary Video 1.}
    \label{fig:sim}
\end{figure}

To assess the robustness of our vision-based control policy in managing complex interactions and avoiding collisions in densely populated environments, we conducted experiments in simulation across four distinct environments designed to challenge navigation and collision avoidance. All simulation experiments were conducted in Isaac Gym using the Unitree Go2 robot URDF model~\footnote{\url{https://github.com/Unitree-Go2-Robot/go2_description}}.
As illustrated in Figure~\ref{fig:sim}, these scenarios included: (A) robots navigating intersecting paths within a dense forest, (B) robots forming a line to traverse a narrow grove path, (C) robots coordinating a bridge crossing from opposing directions, and (D) robots exiting a fenced enclosure through a single narrow opening.

In the forest crossing scenario (A), three groups of robots navigated through a dense forest from different directions: the red group moved to the right, the yellow group moved straight ahead, and the blue group moved to the left. This task necessitated dynamic path adjustments to avoid both static obstacles and other robots. The conflicting routes required real-time processing of visual data to effectively circumvent obstacles through speed and direction modifications.
The grove path scenario (B) tested the system's ability to maintain orderly formation and prevent congestion as robots traversed a narrow path. Six robots started on one side of the path with the goal of reaching the other side. The observed behavior demonstrated precise control, crucial for ensuring continuous and smooth movement within a constrained space.
The bridge crossing scenario (C) evaluated coordinated navigation in a tight space, where 8 robots (divided into two groups of four), approaching from opposite directions synchronized their movements to avoid collisions. This scenario highlighted the system's ability to execute complex maneuvers without explicit communication between robots.
The fence escape scenario (D) examined the collective decision-making and exit strategies of a large group of 48 robots navigating through a single opening from random initial positions inside an enclosure.
The robots exhibited efficient collective behavior, effectively avoiding bottlenecks and demonstrating the system's ability to manage high-density populations through real-time adjustments.

These simulation results underscore the robustness and adaptability of our vision-based control policy,
enabling efficient navigation in challenging, high-density environments. The emergent behaviors, including synchronized movements and efficient exit strategies, demonstrate the system's capacity to
simultaneously avoid static obstacles, respond to the motions of nearby robots, and resolve route conflicts in crowded and cluttered environments.

\subsection*{Scalable vision-based control for large robot swarms}

\begin{figure}[htbp!]
    \centering
    \includegraphics[width=\linewidth]{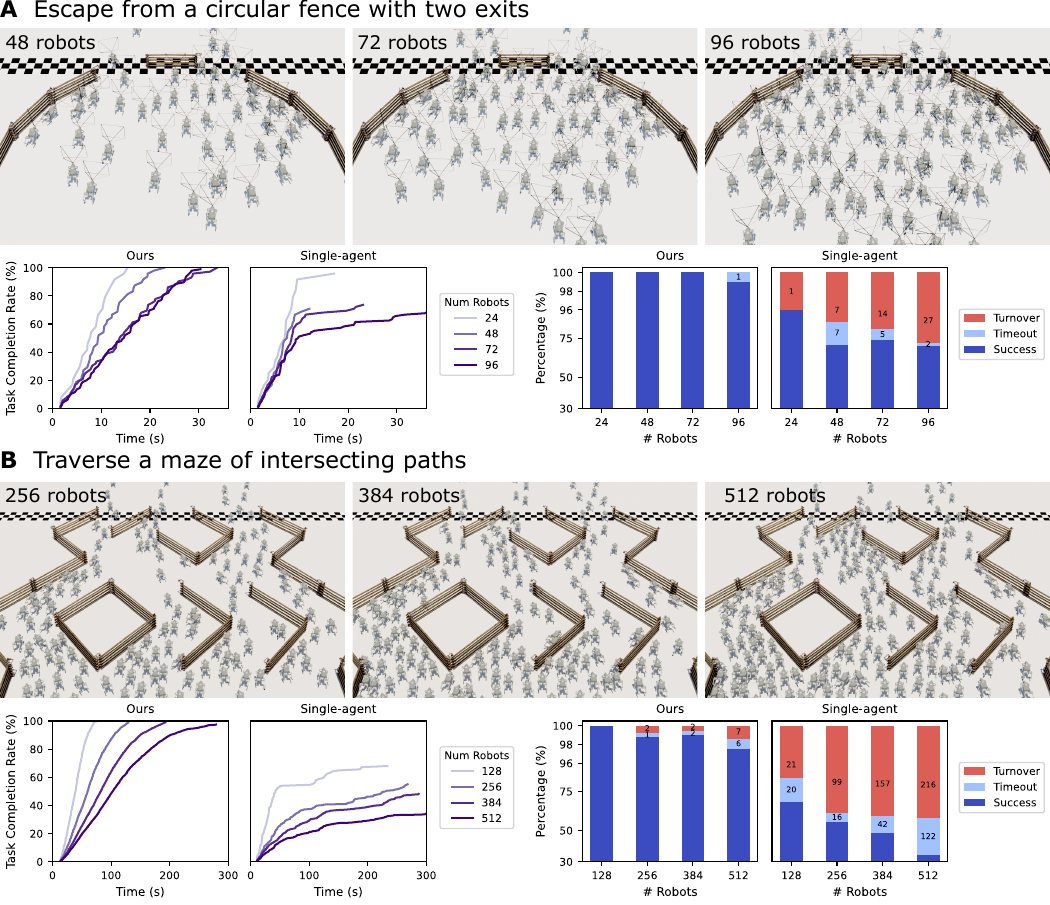}
    \caption{\textbf{Scalability analysis of vision-based swarm navigation.} {\color{black}{\color{black}(A) Two-exit circular fence with 24-96 robots; } (B) intersecting-path maze with 128-512 robots. Curves show the percentage of robots that completed the task over time {\color{black}for our policy and a single-agent policy deployed in the same multi-robot tests}; bars summarize final outcomes (success/timeout/turnover), where {\color{black}\emph{timeout} means that a robot does not finish within the episode horizon and \emph{turnover} means that the robot rolls or pitches beyond 0.5 rad.} } {\color{black}These experiments demonstrate consistent performance across different swarm sizes, with reliable traversal of narrow openings and minimal collisions or congestion.}}
    \label{fig:scale}
\end{figure}

To quantitatively assess the scalability of our vision-based navigation system with increasing robot counts, we performed a series of controlled experiments in two different environments: a circular fence and a complex maze with intersecting paths, as depicted in Figure~\ref{fig:scale}. To further evaluate whether single-robot training transfers to dense multi-robot deployment, we also compare against an end-to-end vision policy trained only in single-agent environments and evaluated in the same large-scale scenes.
We pushed the number of simulated robots to the maximum our desktop computer could handle, aiming to evaluate the system's performance under extreme load.
We employed the task progress rate, defined as the percentage of robots successfully reaching the finish line, as a quantitative metric to track performance over the duration of the experiments.

In the circular fence scenario (Figure~\ref{fig:scale}A), we randomly place robots inside a fenced area with a diameter of \SI{11.8}{\meter} that has two \SI{2}{\meter} exits on one side. The robots must efficiently navigate through these exits to leave the fenced area within \SI{36}{\second}. We test this setup with 24, 48, 72, and 96 robots while measuring how many successfully exit through the fence and cross the finish line (marked with black and white stripes). The robots are initially placed at a random position inside the fence, and the initial positions of any two robots are no less than \SI{0.7}{\meter} apart.
The results show that all robots successfully exit when there are fewer than 96 robots. However, we observe that the robots located between the two exits may be trapped in the middle due to overcrowding, as they are squeezed by robots from both sides. As a result, they are unable to observe the exits on either side and remain stuck. In this dense scenario, one robot failed to exit within the time limit in the test with 96 robots.
Despite this challenge, the system demonstrates great scalability.
96 robots take less than twice the time to exit compared to 48 robots, and similarly, 48 robots take less than twice the time compared to 24 robots. This consistent passage efficiency across different group sizes indicates effective scalability to large swarm navigation through only visual guidance. In the same tests, the single-agent baseline reaches lower final completion rates and produces more turnover events.

In a maze scenario (Figure~\ref{fig:scale}B), robots navigate through fence-enclosed intersecting paths in a confined space. The robots start on one side and must reach a finish line on the opposite side within \SI{300}{\second}. We test this setup with larger groups of 128, 256, 384, and 512 robots.
Results show most robots successfully passed the finish line. Time to reach 75\% completion increases proportionally with the number of robots, demonstrating consistent passage efficiency even with larger groups.
While all robots succeed in the 128-robot scenario, larger groups experience minor failures (less than 3\%) due to two main issues. First, robots sometimes suddenly retreat when they detect other robots entering their field of vision, leading to collisions and turnover events. Second, robots occasionally get their limbs caught in the fence, resulting in a timeout. In the same maze tests, the single-agent baseline shows lower completion rates, especially as the swarm size increases.

\subsection*{Learned multi-robot navigation behaviors}
\label{sec:behaviors}

\begin{figure}[htbp!]
    \centering
    \includegraphics[width=\linewidth]{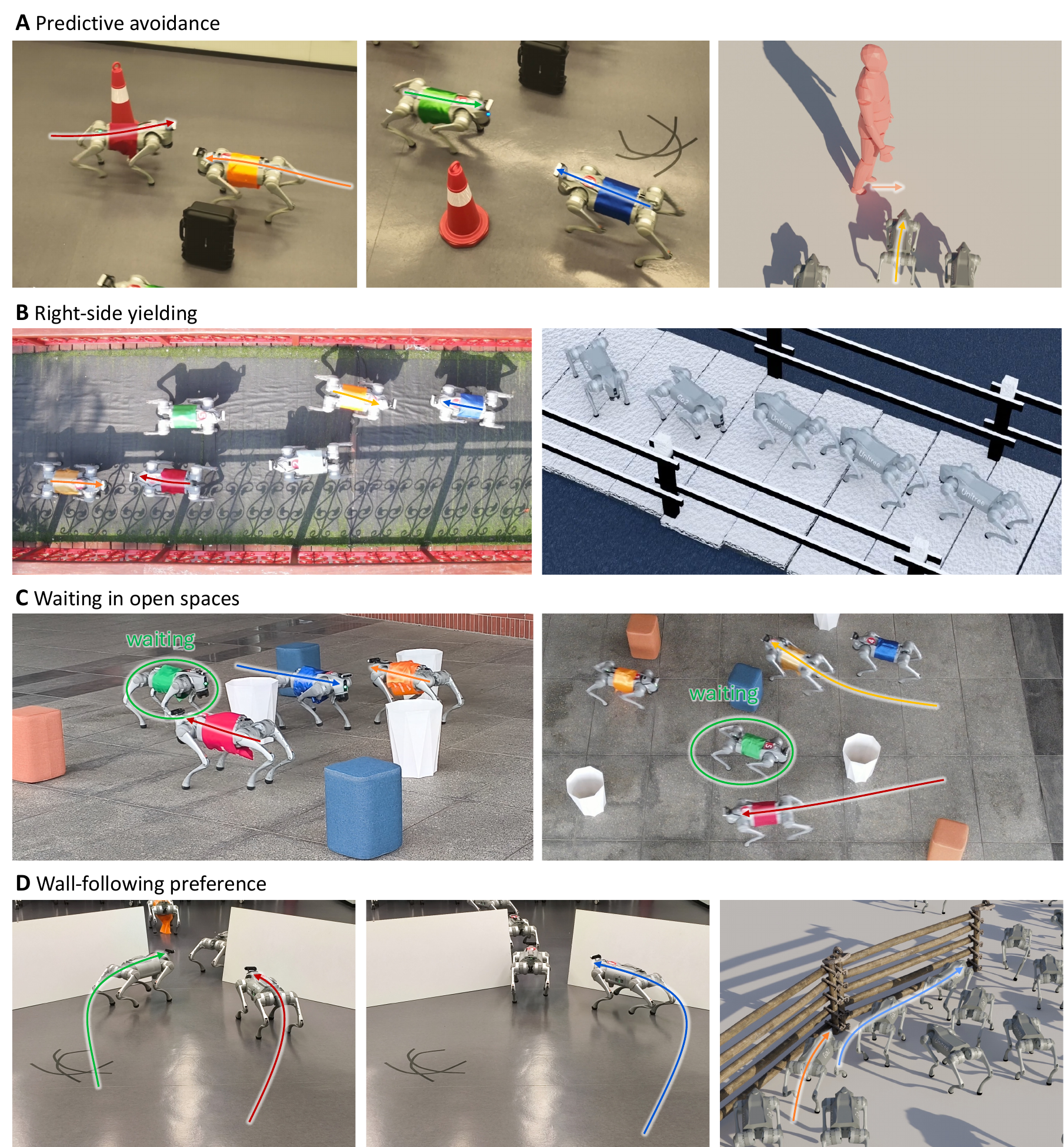}
    \caption{\textbf{Multi-robot navigation behaviors.} Videos are available in Supplementary Video 1. (A) Predict others' movements and avoid collisions. (B) Yield right to prevent deadlocks. (C) Pausing in open areas for narrow passages. (D) Follow walls to avoid unseen side collisions.}
    \label{fig:emerged}
\end{figure}

\paragraph{Reactive responses to nearby agents and obstacles}
The robots demonstrate the ability to
adjust their motions in response to the movements of other robots and dynamic obstacles in their path. This is evident in several scenarios (Figure~\ref{fig:emerged}A): when two robots cross paths in an obstacle-rich environment,
one robot pauses or slightly adjusts its path, increasing space for the other to pass;
when encountering slow-moving obstacles like pedestrians, the robots adapt their avoidance path based on the observed obstacle's direction; and in queuing situations, robots maintain tight formations by
continuing forward once robots ahead are moving. These behaviors,
while primarily local and reactive,
enable smooth navigation and efficient swarm movement with minimal collisions.

\paragraph{Yielding to resolve conflicts}
When two groups of robots approach each other, potential conflicts naturally arise. Interestingly, we observed that robots consistently resolve these encounters by yielding to the right, even for those positioned at the tail end of their groups. Importantly, each robot knows only its own goal in an egocentric coordinate frame and perceives the world solely through a forward-facing depth camera, without access to other robots’ goals, states, or intended actions. The training objective contains no terms encouraging robots to follow peers, coordinate turns, or align trajectories. Consequently, the observed “yielding right” behavior cannot be explained by shared target directions; instead, it emerges from local sensing and repeated multi-robot interactions during training.
{\color{black}We hypothesize that this pattern may reduce collision risk and improve progress in crowded encounters, leading to a stable, self-organized interaction pattern.
This behavior makes robot motions more predictable and reduces the risk of head-on and side collisions. } {\color{black}For example, in the narrow bridge scenario (Figure~\ref{fig:emerged}B), robots from both sides turn right, allowing them to resolve route conflicts and pass safely through the constrained space.}
\paragraph{Pausing in open areas for conflict passages}
{\color{black}In crowded scenarios, robots often stop in open regions before entering conflict passages, thereby reducing the likelihood of blocking others. }
In crossing scenarios with obstacles (Figure~\ref{fig:emerged}C), if both the left and right paths are occupied by other robots, the robot pauses in the open area until the passages become available.
After one robot passes and the path becomes unobstructed, the others move forward based on their reactive visual policy. This reactive behavior allows robots to pass through narrow passages without collisions in our experiments.
\paragraph{Wall-following tendencies under limited perception}
An interesting observation is that the robots exhibit a tendency to move along walls or obstacle edges when navigating complex environments. For instance, when passing through fence exits (Figure~\ref{fig:emerged}D),
robots frequently make contact with the wall due to their limited field of view and reactive vision-based policy. After contact, they tend to move tangentially along the wall surface until the exit becomes visible. This interaction produces a pattern resembling wall-following, observed consistently in simulation and real-world trials. Rather than indicating a deliberate strategy, this behavior arises from the combination of local sensing constraints and reactive collision response. While this sometimes results in temporary congestion near walls, it nonetheless allows robots to eventually traverse the narrow opening without explicit global coordination.

These observed patterns - including pausing, yielding, queuing, and wall-following - are qualitatively similar to behaviors commonly discussed in swarm robotics. The novelty of our work does not lie in the behaviors themselves, but in how they arise: through end-to-end training with vision-based local sensing, a limited field of view, no explicit communication, and the complex dynamics and interactions of quadruped robots. This distinguishes our system from prior approaches based on simplified robots~\cite{mondada2009puck,rubenstein2014programmable, mondada1999development} or idealized settings~\cite{duarte2016evolution, vallegra2018gradual}. Although these behaviors arise from training rather than explicit programming, they are consistent with patterns commonly observed in natural swarms and human crowds. In our system, such behaviors appear spontaneously without centralized control or pre-specified interaction rules, illustrating how local reactive policies can still give rise to coherent multi-robot navigation in cluttered and high-density scenarios.

\subsection*{Behavior comparisons}

\begin{figure}[htbp!]
    \centering
    \includegraphics[width=1.0\linewidth]{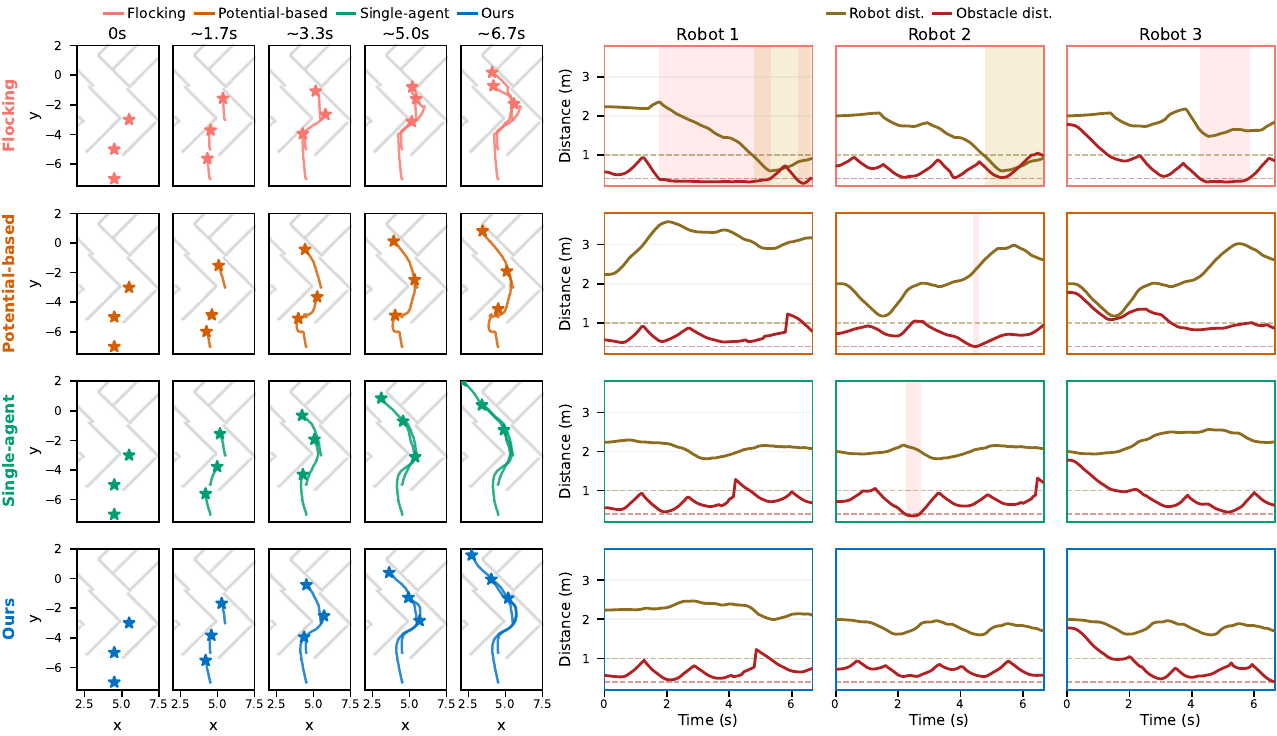}
     \caption{\textbf{Behavioral comparison with heuristic and learning-based baselines.} Left: trajectory snapshots of three robots moving through a cornered passage under the standard flocking controller, the depth-binning potential-based controller, the single-agent policy, and our method. Right: minimum distances from each representative robot to the nearest robot (brown) and nearest obstacle (red) over time. Dashed horizontal lines indicate the safety thresholds of \SI{1.0}{\meter} for robot distance and \SI{0.4}{\meter} for obstacle distance. In the comparison shown here, our policy shows smoother motion and more consistent clearance than the baselines.}
    \label{fig:behavior_cmp}
\end{figure}

We further compare the behaviors produced by our method with three representative baselines that cover both manually designed and learning-based alternatives. The first baseline is the standard flocking algorithm~\cite{reynolds1987flocks}, augmented with target attraction and obstacle repulsion for goal-directed navigation. The second baseline is a depth-binning potential-based controller: the front-facing depth image is manually divided into angular bins, and the closest depth in each bin is converted into repulsive forces that modify the goal-directed velocity command. The third baseline is an end-to-end vision policy trained only in single-agent environments and then evaluated in the same multi-robot scene. All methods are evaluated with identical initial and target configurations. Representative trajectories and distance statistics are shown in Figure~\ref{fig:behavior_cmp}.

As shown in Figure~\ref{fig:behavior_cmp}, the flocking controller shows less stable interactions when goal tracking, obstacle avoidance, and robot-robot avoidance must be handled simultaneously, and the corresponding distance curves indicate more frequent reductions in clearance to nearby robots or obstacles. The potential-based controller reacts to nearby structure from the depth measurements, but its manually designed repulsive fields often induce conservative detours and less consistent passage. The single-agent policy transfers basic visual obstacle-avoidance behavior to the multi-robot scene, but it does not show the same degree of ordered passage or similarly consistent and compact inter-robot spacing as our multi-agent trained policy. By comparison, our policy yields smoother trajectories and more consistent distances to both neighboring robots and obstacles.

These behavior comparisons highlight that, for our method, exposure to multi-robot interactions during training is associated with more ordered passage and more consistent inter-robot spacing than handcrafted depth-based heuristics or single-agent transfer alone. The manually designed baselines rely either on explicit neighbor states or on handcrafted depth-bin rules, whereas the single-agent baseline uses the same visual modality but is not exposed to multi-robot interactions during training. Our policy uses only raw depth images and each robot's own goal information, yet exhibits more stable local coordination in the comparison shown here.

\subsection*{Baseline comparisons}
In this section, we compare the differentiable physics learning (DPL) approach with the traditional reinforcement learning method, PPO~\cite{schulman2017proximal}.
We focus on four different aspects: the efficiency of training samples, the passage efficiency under staggered paths, the scalability to more robots, and the controllability for faster passage.
For every scenario, the simulation is run until all robots reach their goals or a fixed time horizon is reached.

\begin{figure}[htbp!]
    \centering
    \includegraphics[width=\linewidth]{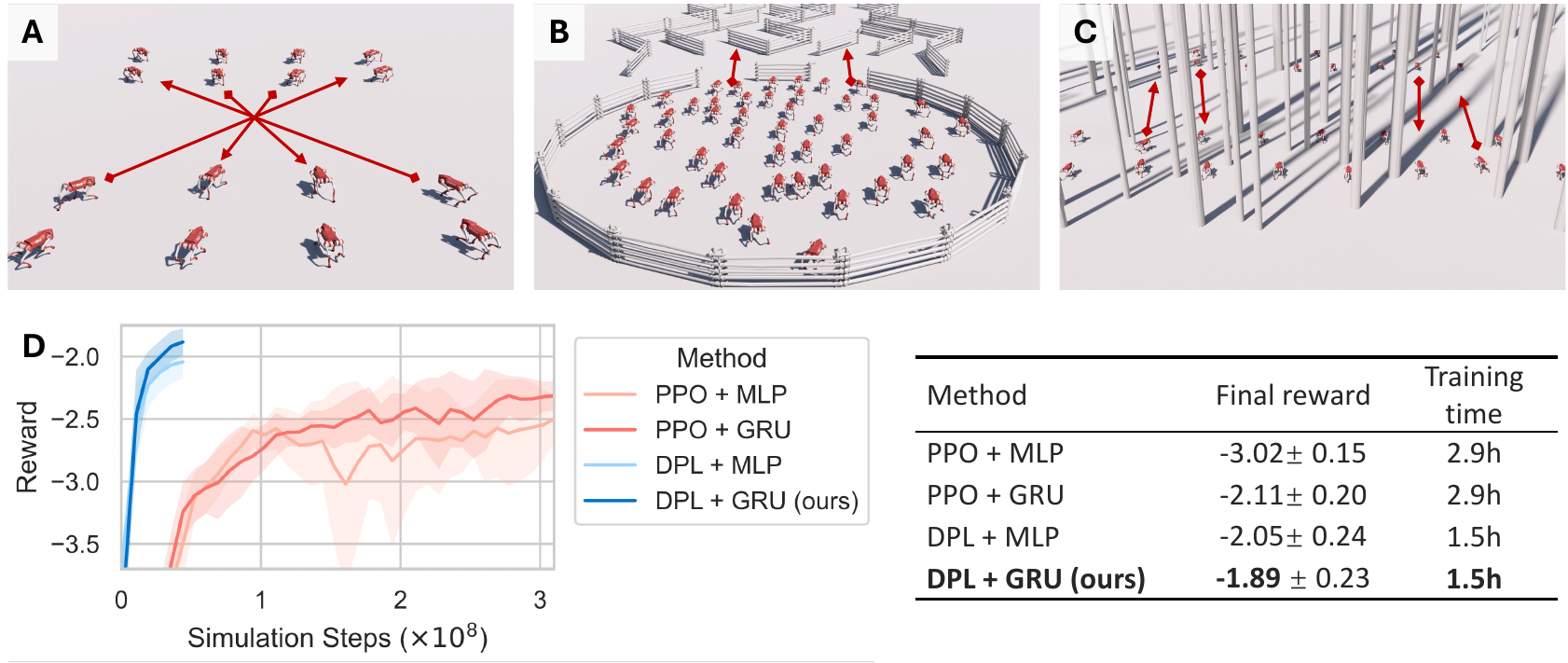}
    \caption{{\bf Baseline comparison against PPO.} (A-C) Three training scenarios: (A) position exchange, (B) exiting a circular fence with radial paths, and (C) position exchange with cylindrical obstacles. (D) Learning curves and statistics of final reward (mean$\pm$std) and training time show that DPL converges with fewer simulation steps and shorter wall-clock time while achieving a higher final reward than PPO. All results are averaged over 5 random seeds (solid line: mean; shaded region: standard deviation).}

    \label{fig:rl_cmp}
\end{figure}

\begin{figure}[htbp!]
    \centering
    \includegraphics[width=1.0\linewidth]{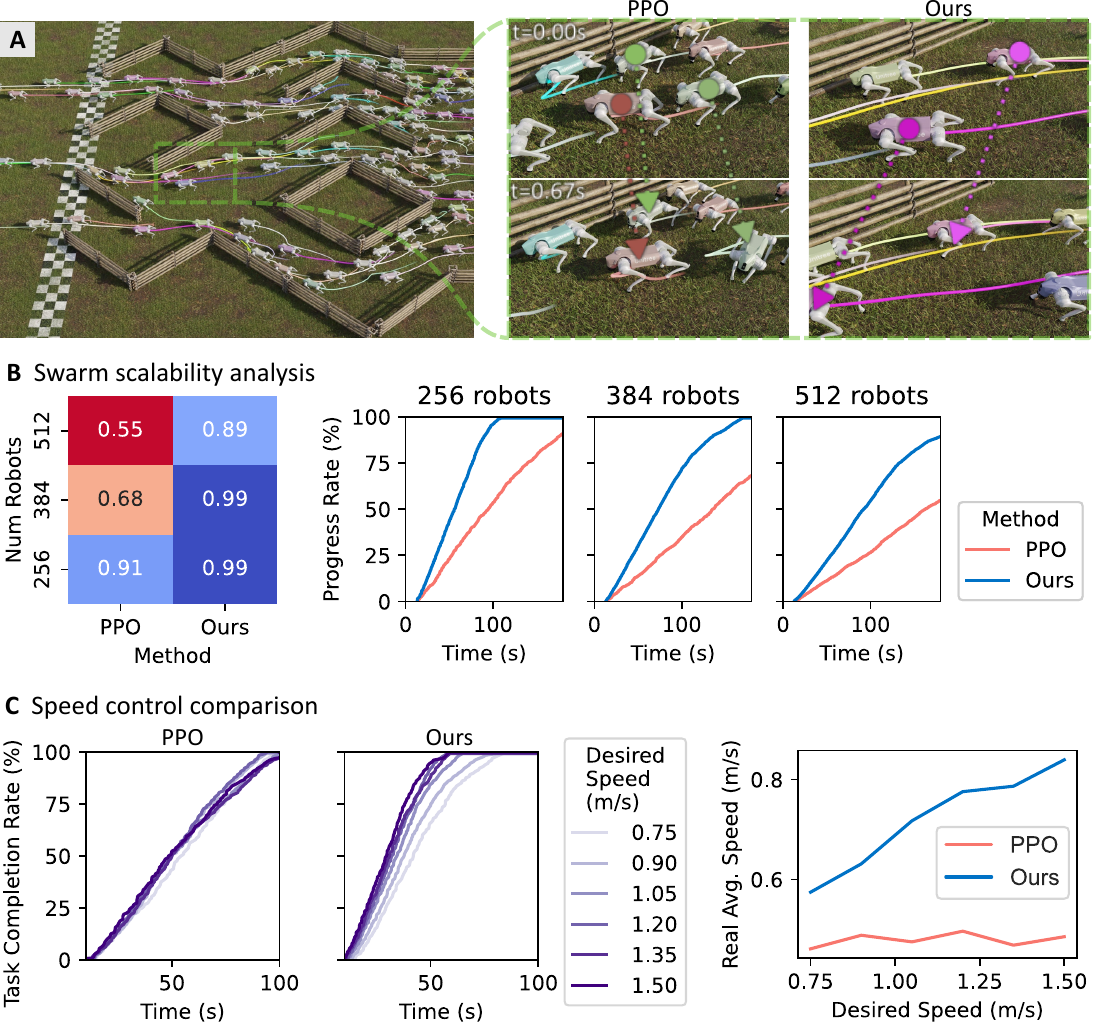}
    \caption{{\bf Scalability to large swarms and speed control ability comparison to PPO.} (A) In intersecting paths, differentiable physics shows a more ordered lineup, while PPO shows more retreats. (B) Scalability under this setting: the heatmap summarizes final task completion rate for each swarm size, and the time plots show completion rate trajectories for 256/384/512 robots, demonstrating consistently faster throughput than PPO as density increases.
    (C)
    Speed controllability: as the commanded speed increases from \SI{0.75}{} to \SI{1.5}{\meter\per\second}, our policy achieves higher completion rates and the realized average speed tracks the command, while PPO shows limited response and saturates at a lower realized speed.
    }
    \label{fig:rl_cmp2}
\end{figure}

\begin{figure}[htbp!]
    \centering
    \includegraphics[width=1.0\linewidth]{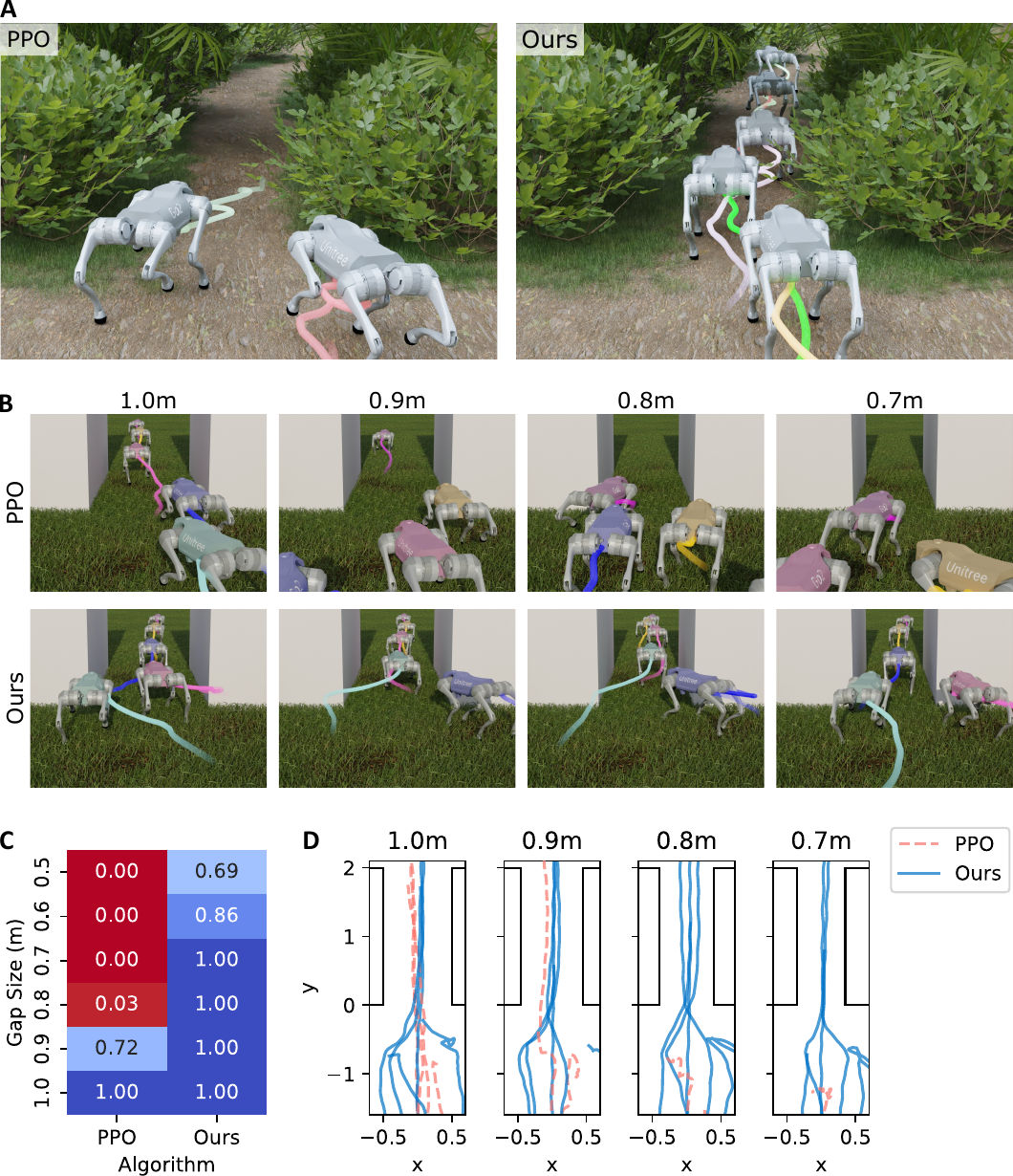}
    \caption{{\bf Narrow-region traversal: PPO vs.\ differentiable physics.} (A) In the bush environment, PPO policy often fails to enter the passage, while the differentiable physics policy executes stable entry and traversal. (B) Qualitative rollouts at decreasing gap widths show that PPO hesitates as the gap narrows, whereas the differentiable physics policy maintains orderly passage. (C) Success rate versus gap width, showing that our method maintains higher success at smaller gaps. (D) The trajectories show that as the passage narrows, our policy continues to enter the channel, while PPO stays outside.}
    \label{fig:narrow}
\end{figure}

First, to compare the training efficiency, we use identical experimental conditions for all evaluated methods: the same training environment, pretrained locomotion policy, and reward function.
The training environment consists of three distinct scenarios illustrated in Figure~\ref{fig:rl_cmp} A-C: (A) 16 robots perform a diagonal position exchange in an \SI{8}{\meter} by \SI{12}{\meter} area with predefined initial and final positions, (B) 48 robots {\color{black}with randomly assigned initial positions} exit an enclosure and navigate through paths formed by fences {\color{black}at a fixed target velocity}, and (C) 32 robots move in opposite directions through cylindrical obstacles {\color{black}between two goal regions placed on either side of the obstacle field. The timeout for each rollout is set to \SI{20}{\second}}. Figure~\ref{fig:rl_cmp}D shows the rewards obtained by both methods across simulation steps. The differentiable simulation approach demonstrates superior sample efficiency in optimization, achieving PPO's final reward with only 2\% of the samples required by PPO.
Moreover, it achieves a higher reward at convergence compared to the PPO~\cite{schulman2017proximal} baseline with fewer training samples. This superior performance can be attributed to the more efficient optimization enabled by the combination of our surrogate model and differentiable physics.

Second, to evaluate the multi-robot navigation performance, we compare the policy optimized by the two methods in the intersecting paths environment. Figure~\ref{fig:rl_cmp2}A qualitatively shows the experimental environment and avoidance behavior of the two methods. In this setup, the robots are initialized on the far right side of the environment, as shown in the figure, with eight robots lined up in a row and a spacing of 1.5 meters between rows. The target velocity is fixed and directed toward the left. The DPL policy shows more continuous and orderly swarm navigation through fences. By contrast, the PPO policy shows an inefficient pattern of abrupt retreats.
Furthermore, Figure~\ref{fig:rl_cmp2}B quantifies this difference. By comparing the task progress rates with the number of robots ranging from 256 to 512, we observe that differentiable physics shows faster task progress and a higher success rate. This can be attributed to its higher efficiency to learn coordination behaviors through first-order gradients.

Third, we evaluate the policies' ability to control navigation speed by varying the commanded velocity from \SI{0.75}{\meter\per\second} to \SI{1.5}{\meter\per\second}. Figure~\ref{fig:rl_cmp2}C presents the task progress rate and average speeds at various speed commands. The DPL policy
demonstrates effective speed scaling, with the actual swarm velocity increasing as the commanded value scales up. In contrast, the PPO policy shows limited speed adaptability - the actual swarm velocity does not respond to higher speed commands.
Our method both scales better to hundreds of robots (higher/faster completion) and maintains controllable speed tracking, whereas PPO degrades with density and fails to increase realized speed under higher commands.
This difference in speed control capability can be attributed to how each method handles multiple competing objectives. As velocity increases, the risk of collision also increases. The DPL approach,
through differentiation, can backpropagate gradients for both speed matching and collision avoidance objectives separately to individual agent actions. In contrast, PPO relies on optimizing the summed reward signal across all objectives. Given its already limited collision avoidance capability, this makes it more challenging to effectively learn speed control and scaling to higher speeds.

Last, we examine the policies' performance in navigating narrow passages.
In the simulated bush environment shown in Figure~\ref{fig:narrow}A, we observed that the policy trained with differentiable physics successfully navigated through narrow paths, while the PPO-trained policy exhibited random exploration outside the path without successfully entering the narrow passage.
To quantitatively evaluate this behavior, we designed a comparative experiment. On a flat terrain, we created a narrow passage using two cubic obstacles. Five robots are lined up vertically in front of the narrow passage, spaced 1.5 meters apart. The target velocity points toward the other side of the passage. We compared both methods by measuring the success rate of robots passing through passage five in a line at different passage widths varying from \SI{1}{\meter} to \SI{0.5}{\meter}.
As shown in Figure~\ref{fig:narrow}B and Figure~\ref{fig:narrow}D, the PPO policy showed difficulty in passage when the gap width was below \SI{0.8}{\meter}, while our policy successfully navigated through gaps ranging from \SI{1.0}{\meter} to \SI{0.7}{\meter}.
Figure~\ref{fig:narrow}C quantifies this performance gap: the PPO policy's effectiveness deteriorated at gaps below \SI{0.8}{\meter}, whereas the differentiable physics policy maintained functionality down to gaps of \SI{0.5}{\meter}.
In tight gaps, our method shows higher success rates than PPO, indicating finer control in constrained multi-robot interactions. The performance gap can be attributed to the differences in policy gradient computation. Differentiable physics leverages deterministic gradients from the physical system to guide policy updates, enabling precise trajectory adjustments. This is critical for navigating narrow passages where centimeter-level deviations could cause failure. In contrast, PPO's reliance on stochastic action sampling introduces unintended variance. This inherent noise in policy training disrupts the fine-grained control required to maintain alignment within tight spaces, leading to higher difficulty in learning precise control.

These comparisons demonstrate the advantages of differentiable physics learning with surrogate models in terms of training efficiency, navigation performance, and adaptability, making it promising for vision-based multi-robot navigation tasks.

\section*{DISCUSSION}

This paper introduces an end-to-end training system for vision-based swarm control of quadruped robots.
Our policy enables decentralized coordination among multiple robots, relying on visual perception to navigate complex and dynamic environments without explicit inter-robot communication. Experimental results demonstrate that our method effectively scales to large swarms, preserving high navigation performance in both simulated and real-world settings. Notably, the learned navigation behaviors—such as trajectory prediction, right-side yielding, pausing in open spaces, and wall-following—resemble patterns observed in biological systems, highlighting the practical applicability of our approach.

The surrogate models are central to our training framework. For the navigation policy network, a point-mass model serves as the surrogate, capturing the essential translational dynamics of the robot's base. This simplification enables efficient gradient-based optimization of the navigation policy. For the locomotion policy, we adopt a rigid-body model to approximate the interactions between the robot's body and the ground, focusing on rotational and translational dynamics caused by ground reaction forces (GRF). These surrogate models balance computational efficiency with physical accuracy during policy training.
Our comparison with traditional reinforcement learning methods reveals advantages in sample efficiency and training wall-clock time when using first-order optimization methods over zeroth-order policy gradients. The improved performance in narrow passages and better scalability with speed commands indicates more precise control and wider generalization. These improvements matter for real-world applications where training efficiency and consistent performance are necessary.

The deployment with six Unitree Go2 robots across varied scenarios---from forests to narrow bridges and cluttered rooms---shows that the approach transfers to real systems. The generalization to these environments suggests that the navigation policies are adaptable. Using only depth camera inputs makes the system practical for real-world deployment compared to approaches requiring external sensing or communication infrastructure.
The results with up to 512 robots in simulation indicate potential for large-scale swarm applications. However, limitations appear in cases of physical interactions and dead-end scenarios in maze environments. These challenges point to areas for future work, such as improving robustness to physical contact and enhancing local decision-making.

Several important challenges remain for future work. First, although the current implementation successfully operates using depth camera inputs, extending the system to handle RGB inputs could enable tasks requiring deeper semantic understanding.
Second, exploring the integration of explicit communication channels could enhance coordination among agents in tasks where explicit communication is beneficial, while still preserving the system's capacity to function effectively without communication.
Finally, exploring policy transfer to robots with different morphologies is a promising direction for increasing the system’s versatility and robustness.

\newpage

\section*{MATERIALS AND METHODS}

\subsection*{Method Overview}
\begin{figure}[htp]
    \centering
    \includegraphics[width=1\textwidth]{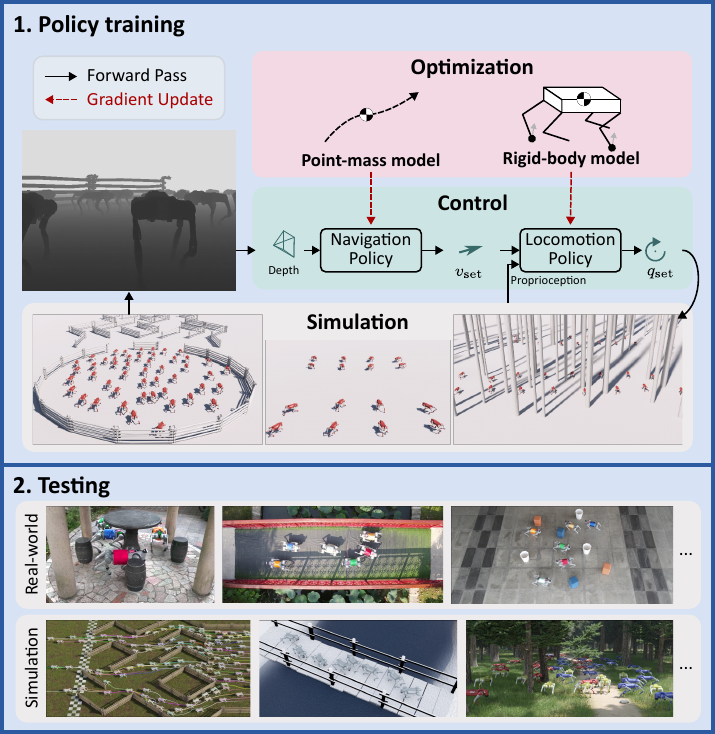}
    \caption{\textbf{Method overview.} Our approach employs hierarchical control of navigation and locomotion policy, optimized via differentiable physics with surrogate models including point-mass model and rigid-body model.}
    \label{fig:method_overview}
\end{figure}

Figure~\ref{fig:method_overview} provides an overview of our system.
We address the problem of quadruped robot navigation, where the robot perceives its environment using a depth camera and navigates toward a target location by following a reference velocity.
The robot learns two hierarchical policies: a high-level navigation policy and a low-level locomotion policy.
The navigation policy processes the depth image to output velocity commands that guide the robot while avoiding obstacles and other agents.
The locomotion policy then translates the velocity commands into desired joint positions for the quadruped's legs, enabling it to execute the motion.
Our decoupled hierarchical design allows the navigation policy to be compatible with various locomotion controllers, providing flexibility in deployment while maintaining robust performance.
A key aspect of our framework is that the robot does not have direct access to information about the state of other robots or agents in its surroundings; it must infer their presence and movements solely from its depth camera observations.
We formulate this task as a hierarchical policy learning problem.

\paragraph{Problem Formulation}
The high-level navigation policy $\pinav{\theta_{\text{nav}}}$ computes a velocity command $v_\text{cmd} \in \mathbb{R}^3$ to follow the reference velocity $v_\text{ref}$
while avoiding obstacles and other agents. Given the depth camera observation $o_t$ and current robot velocity $v_t$, the navigation policy outputs a desired velocity command for the low-level locomotion policy
$$
v_\text{cmd} = \pinav{o_t, v_t; \theta_{\text{nav}}}.
$$
The low-level locomotion policy $\piloc{\theta_\text{loc}}$ receives the velocity command $v_\text{cmd}$ from the navigation policy and computes the desired joint positions $q_\text{desired} \in \mathbb{R}^n$, where n is the number of joints in the quadruped. The locomotion policy aims to track the velocity command as accurately as possible:
$$
q_\text{desired} = \pi_\text{loc}(v_\text{cmd}, q_t; \theta_\text{loc}).
$$

During policy training, the robot receives a navigation cost $l_\text{nav}(x_t, v_\text{cmd})$ and a locomotion cost $l_\text{loc}(x_t, q_\text{desired})$ at each simulation time step $t$.
The objective is to optimize the navigation policy $\pi_\text{nav}$ and the locomotion policy simultaneously by minimizing the following loss functions
\begin{align}
\min_{\pi_\text{nav}} \mathcal{L}(\theta_\text{nav}) &= \mathbb{E} \left[\sum_{t=0}^T l_\text{nav}(x_t, u_t)\right]
= \mathbb{E} \left[\sum_{t=0}^T l_\text{nav}(x_t, \pinav{o_t, v_t} )\right] \\
\min_{\pi_\text{loc}} \mathcal{L}(\theta_\text{loc}) &= \mathbb{E} \left[\sum_{t=0}^T l_\text{loc}(x_t, u_t)\right]
= \mathbb{E} \left[\sum_{t=0}^T l_\text{loc}(x_t, \piloc{o_t, v_t} )\right]
\end{align}
For the definitions of the loss functions, please refer to the \emph{Training Details} section in Appendix.

\subsection*{Policy Optimization via Differentiable Simulation}
This section provides background information about first-order policy optimization via differentiable simulation.
In general, the robot is modeled as a discrete-time dynamical system, characterized by continuous state
and control input spaces, denoted as \( \mathcal{X}  \) and \( \mathcal{U} \),
respectively. The system state is \( x_t \in
\mathcal{X} \) and the control input is \( u_t \in \mathcal{U}
\).
The system dynamics are governed by the function \( f: \mathcal{X} \times
\mathcal{U} \rightarrow \mathcal{X} \), which describes the time-discretized
evolution of the system \( x_{t+1} = f(x_t, u_t) \).
At each time step \( t \), the robot receives a loss signal \( l(x_t, u_t)\), indicating how good the current state and action are.
The control policy is a differentiable function,
such as a neural network,~\( u_t = \pi_{\theta} (o_t)\). The neural network takes the observation \( o_t \) as input and outputs the control input \( u_t \).
The total loss $\mathcal{L}(\theta)$ is to be minimized to find the optimal policy parameters
\(\theta^{\ast}\) by means of gradient descent
\begin{align*}
\theta^{\ast} &= \arg \min_{\theta} \mathcal{L}(\theta) \\
\theta_{k+1} & \leftarrow \theta_k - \alpha \nabla_{\theta} \mathcal{L}(\theta_k),
\end{align*}
where $\alpha$ is the learning rate.
\emph{How can we derive the policy gradient $\nabla_{\theta} \mathcal{L}(\theta_k)$ analytically?}

\paragraph{Surrogate Model}
\label{sec:surrogate_model_optim}
In this section, we generalize the optimization process for a general robot model \( f(x, u) \) with state \( x \) and control input \( u \). The goal is to derive the gradient of the loss function with respect to the policy parameters \( \theta \).
To address the complexity of physical interactions, we use a differentiable surrogate model \( \hat{f}(x, u) \) to approximate the true system dynamics. The surrogate model describes the time-discretized evolution of the system state:
$$
x_{k+1} = \hat{f}(x_k, u_k) + \delta_k = f(x_k, u_k),
$$
where \( \delta_k \) is a slack variable representing the residual components that account for the differences between the surrogate model and the actual system dynamics.
Specifically, we ignore the influence of the model weights on the slack variable, i.e., \( \partial \delta_k / \partial \theta = 0 \).

\paragraph{Gradient Calculation}
Recall that our main goal is to compute the gradient of the loss function with respect to the policy parameters \( \theta \)
\[ \frac{\partial \mathcal{L}_k}{\partial \theta} = \frac{\partial \mathcal{L}}{\partial x_k} \cdot \frac{\partial x_k}{\partial \theta} + \frac{\partial \mathcal{L}}{\partial u_k} \cdot \frac{\partial u_k}{\partial \theta} \]

\paragraph{State Gradient}
The state \( x_k \) depends on the previous state \( x_{k-1} \) and control input \( u_{k-1} \):
\[ x_k = \hat{f}(x_{k-1}, u_{k-1}) + \delta_{k-1}. \]
Thus, the gradient of the state with respect to the policy parameters is
\[ \frac{\partial x_k}{\partial \theta} = \frac{\partial \hat{f}(x_{k-1}, u_{k-1})}{\partial x_{k-1}} \cdot \frac{\partial x_{k-1}}{\partial \theta} + \frac{\partial  \hat{f}(x_{k-1}, u_{k-1})}{\partial u_{k-1}} \cdot \frac{\partial u_{k-1}}{\partial \theta}. \]
We can recursively calculate the state gradient using Backpropagation Through Time (BPTT):
\[
\frac{\partial x_k}{\partial \theta} =
\sum_{i=1}^{k}
\left[
\left( \prod_{j=1}^{i-1} \frac{\partial \hat{f}(x_{k-j}, u_{k-j})}{\partial x_{k-j}} \right) \cdot \frac{\partial \hat{f}(x_{k-i}, u_{k-i})}{\partial u_{k-i}} \cdot \frac{\partial u_{k-i}}{\partial \theta}
\right]
\]

\paragraph{Action Gradient}
The gradient of the action with respect to the policy parameters is
\[ \frac{\partial u_k}{\partial \theta} = \frac{\partial \pi_\theta(o_k)}{\partial \theta}.
\]

\subsection*{Hierarchical Control Training Procedure}

The hierarchical control training procedure is designed to optimize the navigation and locomotion policies within our multi-robot system. This process involves a series of iterative steps that leverage data collected from a simulated environment to refine the policies. The details are in Algorithm~\ref{algo:training}. The training loop begins with the initialization of the navigation policy network \(\pi_{\theta}\) and the locomotion policy network \(\pi_{\phi}\), along with the initial state \(x_0\) and the goal location \(\gamma\).

During each iteration of the training loop, data is collected from the simulation environment using Isaac Gym~\cite{makoviychuk2021isaac}. This data includes depth images, goal velocities, and robot states. The goal velocity is computed from the robot's current position and its goal position. The navigation policy network \(\pi_{\theta}\) processes the depth image and goal velocity to output velocity and yaw commands, which are then fed into the locomotion policy network \(\pi_{\phi}\). The locomotion policy network generates joint angle commands that control the robot's movement.

In deployment, each robot is assigned an independent goal at the beginning, and robots do not exchange states or goals. The navigation policy network receives a goal velocity vector that always points toward the robot’s goal.

The simulation environment, specifically Isaac Gym, steps forward with the joint angle commands, producing the next state \(x_{k+1}\). The navigation loss \(\mathcal{L}_n\) and the locomotion loss \(\mathcal{L}_l\) are computed based on the deviation of the robot's actual state from the desired state. These losses are accumulated over a number of steps \(S_n\) and \(S_l\) for the navigation and locomotion policies, respectively.

Once the losses are accumulated, we compute the gradients of the loss functions with respect to the policy parameters via surrogate models. The gradients are then used to update the policy parameters using the AdamW optimizer~\cite{kingma2014adam}. This process is repeated until the training converges, and the output is the optimized navigation and locomotion policies that enable the multi-robot system to navigate and coordinate effectively.

For real-world deployment, we utilize Unitree robots. The design choice of decoupling navigation policy from the locomotion policy provides the flexibility to also integrate Unitree's built-in locomotion controller.
This choice allowed us to focus on validating our policy's obstacle avoidance and coordination capabilities in complex multi-robot scenarios.
The successful integration with an existing locomotion controller demonstrates the practical advantages of our decoupled hierarchical design, as the navigation policy can be readily deployed with different locomotion solutions while maintaining robust performance.

\begin{algorithm}[]
\caption{Hierarchical Control Training Procedure}
\label{algo:training}
\begin{algorithmic}[1]

\State \textbf{Input:} $S_n$ and $S_l$: gradient accumulation steps for navigation policy and locomotion policy respectively.
\State \textbf{Output:} Optimized navigation and locomotion policies
\State \textbf{Initialize:} Navigation policy network $\pi_{\theta}$, locomotion policy network $\pi_{\phi}$, initial state $x_0$, goal location $\gamma$.

\State $k \gets 1$
\State $L_n \gets 0$ \Comment{Accumulated navigation loss}
\State $L_l \gets 0$ \Comment{Accumulated locomotion loss}

\While{training not converged}
    \State \textbf{Simulation Pass:}
    \State $o_k \gets \text{Depth image \& goal velocity}$
    \State $v_k, \omega_k \gets \pi_{\theta}(o_k)$ \Comment{Navigation policy}
    \State $q_k \gets \pi_{\phi}(v_k, \omega_k, s_k)$ \Comment{Locomotion policy}
    \State $x_{k+1} \gets \text{IsaacGym}(x_k, q_k)$ \Comment{Step simulation}
    \State $L_n \gets L_n + \mathcal{L}_n(x_{k+1}; \gamma)$ \Comment{Navigation loss}
    \State $L_l \gets L_l + \mathcal{L}_l(x_{k+1}; v_k, \omega_k)$ \Comment{Locomotion loss}
    \State $k \gets k + 1$

    \If{$k \mod S_n = 0$}
    \State \textbf{Navigation Policy Optimization:}
        \State $\bar{L}_n \gets {L_n}/{S_n}$ \Comment{Average navigation loss}
        \State $g_{\theta} \gets \text{backward}(\bar{L}_n, f_\text{point mass})$ \Comment{Navigation loss gradient}
        \State $\theta \gets \theta - \eta g_{\theta}$ \Comment{Update navigation policy}
        \State $L_n \gets 0$ \Comment{Reset accumulated navigation loss}
    \EndIf
    \If{$k \mod S_l = 0$}
    \State \textbf{Locomotion Policy Optimization:}
        \State $\bar{L}_l \gets {L_l}/{S_l}$ \Comment{Average locomotion loss}
        \State $g_{\phi} \gets \text{backward}(\bar{L}_l, f_\text{rigid body})$ \Comment{Locomotion loss gradient}
        \State $\phi \gets \phi - \eta g_{\phi}$ \Comment{Update locomotion policy}
        \State $L_l \gets 0$ \Comment{Reset accumulated locomotion loss}
    \EndIf
\EndWhile

\end{algorithmic}
\end{algorithm}

\paragraph{Navigation Policy Optimization}

The navigation policy network in our system leverages a point-mass model to provide accurate gradients for efficient optimization. The point-mass dynamics are described by the discrete-time equation:
\begin{equation}
p_{k+1} = p_k + v_k \cdot \Delta t + r_k,
\end{equation}
where \( p_k \) denotes the robot's position at time step \( k \), \( v_k \) represents the velocity at time step \( k \), and \( \Delta t \) is the integration time step. The velocity \( v_k \) is determined by the navigation policy network, such that \( v_k = \pi_\theta(o_k) \), where \( \pi_\theta \) is the policy parameterized by \(\theta\), and \( o_k \) is the observation at time step \( k \). The term \( r_k \) denotes a residual component that accounts for variations in the robot's motion, such as imperfect locomotion or unmodeled dynamics, which we ignore in the gradient calculation and optimization process.

The training objective is to minimize the loss function \( L_k \) at each time step, defined as:
\begin{equation}
L_k = \mathcal{L}_n(p_k, v_k),
\end{equation}
where \(\mathcal{L}\) quantifies the deviation of the current state from the desired trajectory.

To optimize the policy parameters \(\theta\), we compute the gradient of the loss function with respect to \(\theta\). The chain rule is employed to derive this gradient:
\begin{equation}
\frac{\partial L_k}{\partial \theta} = \frac{\partial \mathcal{L}}{\partial p_k} \cdot \frac{\partial p_k}{\partial \theta} + \frac{\partial \mathcal{L}}{\partial v_k} \cdot \frac{\partial v_k}{\partial \theta}.
\end{equation}

For the sake of simplicity in derivation, we assume without loss of generality that the robot perfectly follows the navigation command, i.e., \( v_k = \pi_\theta(o_k) \). The gradient of the velocity with respect to the policy parameters is straightforward:
\begin{equation}
\frac{\partial v_k}{\partial \theta} = \frac{\partial \pi_\theta(o_k)}{\partial \theta}.
\end{equation}
In our implementation, we incorporate a time delay and exponential moving average for the navigation model outputs to more accurately capture the robot's response.

Since the position at each time step depends recursively on the previous positions and velocities, we express the gradient of the position with respect to the policy parameters in terms of the gradients at the previous time steps. Specifically, the gradient of the position at time step \( k \) is influenced by the velocity at the previous time step \( k-1 \):
\begin{equation}
\frac{\partial p_k}{\partial \theta} = \frac{\partial p_{k-1}}{\partial \theta} + \Delta t \cdot \frac{\partial \pi_\theta(o_{k})}{\partial \theta} + \frac{\partial r_{k}}{\partial \theta}.
\end{equation}

We ignore the influence of policy on model error, which is acceptable for small perturbations. Thus, we obtain:
\begin{equation}
\frac{\partial p_k}{\partial \theta} = \frac{\partial p_{k-1}}{\partial \theta} + \Delta t \cdot \frac{\partial \pi_\theta(o_{k-1})}{\partial \theta}.
\end{equation}

By expanding the recursion, we obtain:
\begin{equation}
\frac{\partial p_k}{\partial \theta} = \sum_{i=1}^{k} \Delta t \cdot \frac{\partial \pi_\theta(o_{i-1})}{\partial \theta}.
\end{equation}

This recursive relationship facilitates the calculation of the position gradient at each time step, allowing us to efficiently compute the overall gradient of the loss function with respect to the policy parameters \(\theta\). By iterating backwards through the time steps, we can perform backpropagation to update the policy parameters, thereby optimizing the navigation policy.

\paragraph{Locomotion Policy Optimization}
The locomotion policy network in our system leverages a rigid-body model to provide accurate gradients for efficient optimization. The rigid-body dynamics are described by the discrete-time equation:
\begin{align*}
\dot{\mathbf{p}}_{WB} &= \mathbf{v}_{WB} &
\dot{\mathbf{v}}_{WB} &= \frac{1}{m} \sum_i \mathbf{f}_i + \mathbf{g} \\
\dot{\mathbf{q}}_{WB} &= \frac{1}{2} \Lambda(\boldsymbol{\omega}_{B}) \cdot \mathbf{q}_{WB} &
\dot{\boldsymbol{\omega}}_{B} &= \mathbf{I}^{-1} \left( \boldsymbol{\eta} - \boldsymbol{\omega}_{B} \times (\mathbf{I}\boldsymbol{\omega}_{B}) \right)
\end{align*}
where \( \mathbf{p}_{WB} \) denotes the robot's position, \( \mathbf{v}_{WB} \) represents the velocity, \( \mathbf{q}_{WB} \) is the orientation represented by a unit quaternion, and \( \boldsymbol{\omega}_{B} \) is the angular velocity. The control inputs are the ground reaction forces \( \mathbf{f}_i \) from the legs.

The state \( x \) consists of \( \mathbf{p}_{WB}, \mathbf{v}_{WB}, \mathbf{q}_{WB} \), and \( \boldsymbol{\omega}_{B} \), while the control input \( u \) consists of the joint angle commands that are converted into ground reaction forces \( \mathbf{f}_i \) via a PD controller and forward dynamics. Specifically, the ground reaction forces \( \mathbf{f}_i \) are computed by solving:
\begin{equation*}
\mathbf{J}_i^T\mathbf{f}_i = \boldsymbol{\tau}_i,
\end{equation*}
where \( \mathbf{J}_i \) is the foot Jacobian matrix, and \( \boldsymbol{\tau}_i \) is the torque applied by the \( i \)-th leg. The torques \( \boldsymbol{\tau} \) are generated by a PD controller that computes the required motor torque based on the difference between the reference joint position \( \mathbf{q}^\text{ref} \) and the current joint position \( \mathbf{q} \), as well as the difference between the reference joint velocity \( \mathbf{\dot{q}}^\text{ref} \) and the current joint velocity \( \mathbf{\dot{q}} \):
\begin{equation*}
\boldsymbol{\tau} = \mathbf{k_p} (\mathbf{q}^\text{ref} - \mathbf{q}) + \mathbf{k_d} (\mathbf{\dot{q}}^\text{ref} - \mathbf{\dot{q}}),
\end{equation*}
where \( \mathbf{k_p} \) and \( \mathbf{k_d} \) are the proportional and derivative gains, respectively.

The training objective is to minimize the loss function \( L_k \) at each time step, defined as:
\begin{equation}
L_k = \mathcal{L}_l(\mathbf{p}_{WB}, \mathbf{v}_{WB}, \mathbf{q}_{WB}, \boldsymbol{\omega}_{B}; \mathbf{f}_i),
\end{equation}
where \(\mathcal{L}_l\) quantifies the deviation of the current state from the desired trajectory.

The backward pass for the locomotion policy is consistent with Section Surrogate Model Optimization. Similar to the navigation policy optimization, the gradient of the loss function with respect to the policy parameters \(\phi\) is computed using the chain rule. The state \( x_k \) at each time step depends on the control inputs $u_{k-1}$ and recursively on the previous state $x_{k-1}$, with the residual term $r_k$ being ignored for simplicity in the gradient calculation. This recursive relationship facilitates the calculation of the state gradient at each time step, enabling efficient computation of the overall gradient of the loss function with respect to the policy parameters \(\phi\).

\section*{Data availability}
The recorded videos for real-world experiments and data in simulated experiments are included in the supplementary material and available at \url{https://drive.google.com/file/d/1gCK_Etd1bTRdJB8tFYdORBU3zJI5xK7a/view}.

\section*{Code availability}
The code for training and evaluation in simulated environments is available at \url{https://drive.google.com/file/d/1Owwn3K3f7btQro0g8nE_bWRay008JnKt/view}.

\section*{Acknowledgments}
The paper is supported in part by the National Natural Science Foundation of China (No.62325109, 62595733, 62561160155).

\newpage
\bibliography{scibib}
\newpage
\begin{appendices}
\renewcommand*{\thefigure}{\arabic{figure}}
\renewcommand*{\thetable}{\arabic{table}}
\renewcommand*{\thesection}{\arabic{section}}
\renewcommand{\appendixname}{Note}
\setcounter{figure}{0}
\setcounter{table}{0}
\setcounter{section}{0}

\newpage
\section*{Supplementary Materials}

\subsection*{Table of Contents}
The supplementary information in this document includes:\\
Supplementary Sections 1-3\\
Supplementary Tables 1-2\\

\noindent
Other supplementary information includes:\\
Supplementary Video 1\\

\section*{Supplementary Tables}

\begin{table}[ht]
\centering
\caption{Training Hyperparameters for Locomotion and Navigation Policies}
\label{tab:training_params}
\begin{tabular}{l|l|l}
\toprule
\textbf{Parameter} & \textbf{Locomotion Policy} & \textbf{Navigation Policy} \\
\midrule
\multicolumn{3}{l}{\textbf{Training Parameters}} \\
\midrule
Robots per environment & 96 & 96 \\
Number of environments & 1 & 4 \\
Maximum simulation time & 20 s & 20 s \\
Simulation frequency & 60 Hz & 60 Hz \\
Network prediction frequency & 60 Hz & 15 Hz \\
Optimization steps & 5,000 & 1,250 \\
Simulation steps per optimization step & 20 & 80 \\
Optimizer & AdamW & AdamW \\
Learning rate & 0.001 & 0.001 \\
Betas & (0.5, 0.95) & (0.5, 0.95) \\
LR scheduler & Cosine annealing & Cosine annealing \\
Minimum learning rate & 0.0001 & 0.00001 \\
\midrule
\multicolumn{3}{l}{\textbf{Motion Parameters}} \\
\midrule
Gait pattern & Trot & - \\
Gait frequency & 1.7 to 2.5 Hz & 1.7 to 2.5 Hz \\
Swing ratio & 0.5 & - \\
Maximum speed & 0.51 to 1.5 m/s & 0.51 to 1.5 m/s \\
\bottomrule
\end{tabular}
\end{table}

\begin{table}
\centering
\caption{Details of Locomotion and Navigation Policy Networks}
\label{tab:network_arch}
\begin{tabularx}{\textwidth}{l|l}
\toprule
\textbf{Policy} & \textbf{Architecture} \\
\midrule
\textbf{Locomotion}
  & \textbf{Observation (dim: 35):} \\
  & \quad $\bullet$ 2 phase variables (sin / cos) \\
  & \quad $\bullet$ 3 target velocity components \\
  & \quad $\bullet$ 1 target yaw \\
  & \quad $\bullet$ 3 base angular velocity (omega) \\
  & \quad $\bullet$ 2 base orientation (roll / pitch) \\
  & \quad $\bullet$ 12 joint angles \\
  & \quad $\bullet$ 12 joint velocities \\
  & \textbf{Action (dim: 12):} Joint angles \\
  & \textbf{Network:} MLP with hidden sizes [256, 256] and ReLU activation \\
\midrule
\textbf{Navigation}
  & \textbf{Observation:} \\
  & \quad $\bullet$ 16×12 pixel image \\
  & \quad $\bullet$ 2 target velocity components \\
  & \quad $\bullet$ 192 GRU hidden state dimensions \\
  & \textbf{Action (dim: 2):} Velocity setpoint \\
  & \textbf{Network Architecture:} \\
  & \quad \textit{Image Processing Branch:} \\
  & \quad \quad $\bullet$ Conv2d (1→32, kernel=2, stride=2) → LeakyReLU \\
  & \quad \quad $\bullet$ Conv2d (32→64, kernel=3) → LeakyReLU \\
  & \quad \quad $\bullet$ Conv2d (64→128, kernel=3) → LeakyReLU \\
  & \quad \quad $\bullet$ Flatten → Linear (1024→192) \\
  & \quad \textit{Observation Processing Branch:} \\
  & \quad \quad $\bullet$ Linear (9→192) with weights scaled by 0.5 \\
  & \quad \textit{Feature Fusion \& Recurrent Processing:} \\
  & \quad \quad $\bullet$ Addition of CNN and observation features \\
  & \quad \quad $\bullet$ GRU (192→192) \\
  & \quad \quad $\bullet$ Linear (192→4) with LeakyReLU activation \\
\bottomrule
\end{tabularx}
\end{table}

\section*{Supplementary Sections}
\section{Training Hyperparameters}
This section provides an overview of the training hyperparameters (see Table~\ref{tab:training_params}), network architectures (see Table~\ref{tab:network_arch}), and loss functions (see Sec.~\ref{sec:nav_loss} and Sec.~\ref{sec:loco_loss}) used in our approach.

\section{Navigation Policy Loss Function}
\label{sec:nav_loss}

The navigation policy loss function is designed to optimize the robot's trajectory towards the target while avoiding obstacles and maintaining smooth motion. The total loss is a weighted sum of several components:
\begin{equation}
\mathcal{L}_n = \alpha_v \mathcal{L}_v + \alpha_{\text{obs}} \mathcal{L}_{\text{obs}} + \alpha_{c} \mathcal{L}_{c} + \alpha_{\text{spd}} \mathcal{L}_{\text{spd}} + \alpha_{\text{acc}} \mathcal{L}_{\text{acc}} + \alpha_{\text{dir}} \mathcal{L}_{\text{dir}} + \alpha_{\text{tgt}} \mathcal{L}_{\text{tgt}} + \alpha_t \mathcal{L}_t
\end{equation}
where \( \alpha_v, \alpha_{\text{obs}}, \alpha_{c}, \alpha_{\text{spd}}, \alpha_{\text{acc}}, \alpha_{\text{dir}}, \alpha_{\text{tgt}}, \alpha_t \) are the coefficients for each loss component.

\subsection{Velocity Loss}

The velocity loss measures the deviation of the planned velocity from the target velocity:
\begin{equation}
\mathcal{L}_v = \|v_{\text{tgt}} - v_p\|^2
\end{equation}
where \( v_{\text{tgt}} \) is the target velocity and \( v_p \) is the planned velocity.

\subsection{Obstacle Avoidance Loss}

The obstacle avoidance loss encourages the robot to maintain a safe distance from obstacles:
\begin{equation}
\mathcal{L}_{\text{obs}} = v_{\text{obs}} \cdot (1 - d_{\text{obs}})_+^2
\end{equation}
where \( d_{\text{obs}} \) is the Euclidean distance to the nearest obstacle, \( v_{\text{obs}} \) is the scalar speed towards the obstacle, and \( (\cdot)_+ \) denotes the ReLU function.

\subsection{Collision Loss}

The collision loss penalizes the robot for getting too close to obstacles:
\begin{equation}
\mathcal{L}_{c} = v_{\text{obs}} \cdot \text{Softplus}(-32 \cdot d_{\text{obs}})
\end{equation}
where \( \text{Softplus}(x) = \log(1 + e^x) \) and \( d_{\text{obs}}, v_{\text{obs}} \) follow the previous definitions.

\subsection{Speed Loss}

The speed loss penalizes the robot for planning velocities that are too high, which can lead to instability:
\begin{equation}
\mathcal{L}_{\text{spd}} = \sum_{i=1}^{N} \|v_{p,i}\|^2
\end{equation}
where \( v_{p,i} \) is the planned velocity at time step \( i \).

\subsection{Acceleration Loss}

The acceleration loss penalizes the robot for sudden changes in velocity, which can cause jerky motion:
\begin{equation}
\mathcal{L}_{\text{acc}} = \sum_{i=1}^{N} \|a_i\|^2
\end{equation}
where \( a_i \) is the acceleration at time step \( i \), calculated as:
\begin{equation}
a_i = \frac{v_{p,i} - v_{p,i-1}}{\Delta t}
\end{equation}
and \( \Delta t \) is the time step duration.

\subsection{Direction Loss}

The direction loss \( \mathcal{L}_{\text{dir}} \) ensures that the planned velocity is aligned with the robot's forward direction:
\begin{equation}
\mathcal{L}_{\text{dir}} = 1 - \cos({\bf f}, v_p)
\end{equation}
where \( \mathbf{f} \) is the robot's forward direction vector.

\subsection{Target Direction Loss}

Likewise, the target direction loss \( \mathcal{L}_{\text{tgt}} \) ensures that the robot's forward direction is aligned with the target velocity. The target direction loss is then calculated as the cosine similarity between the robot's forward direction and the target velocity direction:
\begin{equation}
\mathcal{L}_{\text{tgt}} = 1 - \cos(\mathbf{f}, v_\text{tgt}).
\end{equation}

\subsection{Time Loss}

The time loss \( \mathcal{L}_t \) encourages the robot to move and complete the trajectory in minimal time:
\begin{equation}
\mathcal{L}_t = -\log(\|v_p\| + 10^{-2})
\end{equation}

This loss function ensures that the robot navigates efficiently towards the target while avoiding obstacles and maintaining smooth motion.

\subsection{Coefficient Values}

The coefficients for each loss component in the navigation policy loss function are set as follows:
$
\alpha_v = 1,
\alpha_{\text{obs}} = 2,
\alpha_{c} = 2,
\alpha_{\text{spd}} = 0.01,
\alpha_{\text{acc}} = 0.01,
\alpha_{\text{dir}} = 1,
\alpha_{\text{tgt}} = 0.5,
\alpha_t = 0.2.
$

\section{Locomotion Policy Loss Function}
\label{sec:loco_loss}

The locomotion policy loss function is designed to optimize the robot's movement to follow the planned velocity and yaw rate while maintaining stability and minimizing unnecessary actions. The total loss is a weighted sum of several components:
\begin{equation}
    \begin{aligned}
    \mathcal{L}_l
    = \alpha_v \mathcal{L}_v
    + \alpha_R \mathcal{L}_R
    + \alpha_\omega \mathcal{L}_\omega
    + \alpha_h \mathcal{L}_h
    + \alpha_g \mathcal{L}_g
    + \alpha_q \mathcal{L}_q
    + \alpha_\tau \mathcal{L}_\tau
    + \alpha_s \mathcal{L}_{\text{s}}
    \end{aligned}
\end{equation}
where \( \alpha_v, \alpha_R, \alpha_\omega, \alpha_h, \alpha_g, \alpha_q, \alpha_\tau, \alpha_s \) are the coefficients for each loss component.

\subsection{Velocity Tracking Loss}

The velocity tracking loss measures the deviation of the robot's actual velocity \( \mathbf{v} \) from the reference velocity \( \mathbf{v}^\text{ref} \). The reference velocity at this stage is the output of the navigation policy, which is detached to prevent gradient flow back to the navigation policy. The velocity tracking loss is calculated as follows:
\begin{equation}
\mathcal{L}_v = \| \mathbf{v} - \mathbf{v}^\text{ref}\|^2
\end{equation}

\subsection{Orientation Tracking Loss}

The orientation tracking loss \( \mathcal{L}_R \) measures the deviation of the robot's actual orientation from the reference orientation. The reference orientation is set to be flat with a yaw angle planned by the navigation policy. The orientation tracking loss is calculated as follows:
\begin{equation}
\mathcal{L}_R = 1 - \cos(\theta_{\text{R}})
\end{equation}
where \( \theta_{\text{R}} \) is the angle between the actual and reference orientations.

\subsection{Angular Velocity Tracking Loss}

The angular velocity tracking loss \( \mathcal{L}_\omega \) measures the deviation of the robot's actual angular velocity \( \boldsymbol{\omega} \) from the reference angular velocity \( \boldsymbol{\omega}^\text{ref} \). The reference angular velocity is along the direction of the minimal rotation required to align the robot's orientation with the reference, which is derived from the difference between the current orientation and the reference orientation. The reference angular velocity is calculated as follows:
\begin{equation}
\boldsymbol{\omega}^\text{ref} = \text{AxisAngle}(R^{-1} R^\text{ref})
\end{equation}
where \( R \) is the current orientation quaternion and \( R^\text{ref} \) is the reference orientation quaternion. We then replace the yaw component of the reference angular velocity by the planned yaw rate. The angular velocity tracking loss is calculated as follows:
\begin{equation}
\mathcal{L}_\omega = \| \boldsymbol{\omega} - \boldsymbol{\omega}^\text{ref}\|^2
\end{equation}
where \( \boldsymbol{\omega} \) is the actual angular velocity and \( \boldsymbol{\omega}^\text{ref} \) is the reference angular velocity.

\subsection{Body Height Loss}

The body height loss is designed to ensure the robot maintains a reasonable body height for operation. The body height \( h \) is defined as the average distance from the robot's base position to the stance leg positions along the body upward direction. The body height loss is calculated as follows:
\begin{equation}
\mathcal{L}_h = \| h - h^\text{ref}\|^2
\end{equation}
where \( h \) is the actual body height and \( h^\text{ref} \) is the body height at stance.

\subsection{Gravity Projection Loss}

The gravity projection loss is aimed at maintaining the robot's upright pose by minimizing the horizontal component of the gravity vector in the robot's body frame. The gravity projection loss is calculated as follows:
\begin{equation}
\mathcal{L}_g = \| \mathbf{g}_\text{proj} \|^2
\end{equation}
where \( \mathbf{g}_\text{proj} \) is the projection of the gravity vector onto the horizontal plane of the robot's body.

\subsection{Action Regularization Loss}

The action regularization loss \( \mathcal{L}_q \) penalizes large actuator control targets to prevent excessive movements that may lead to instability or damage. The action regularization loss is calculated as follows:
\begin{equation}
\mathcal{L}_q = \| \mathbf{q}^\text{ref} \|^2
\end{equation}
where \( \mathbf{q}^\text{ref} \) represents the joint motor angle control targets w.r.t the motor positions at stance, which is the direct output of the locomotion policy. This loss ensures that the robot's actions remain within a safe and stable range.

\subsection{Torque Regularization Loss}

The torque regularization loss \( \mathcal{L}_\tau \) is designed to conserve motor torque, thereby making locomotion more energy efficient. The torque regularization loss is calculated as follows:
\begin{equation}
\mathcal{L}_\tau = \| \boldsymbol{\tau} \|^2
\end{equation}
where \( \boldsymbol{\tau} \) represents the motor torques. This loss encourages the robot to use minimal torque while maintaining stable and efficient locomotion.

\subsection{Swing Leg Loss}

The swing leg loss supervises swing legs to follow a heuristic trajectory. The trajectory is calculated by fitting a quadratic polynomial to the starting position, mid-air height, and landing position of each leg. The starting position is the leg position at the beginning of the swing phase, and the landing position is computed using Raibert's heuristic~\cite{raibert1986legged}. The swing leg loss is calculated as follows:
\begin{equation}
\mathcal{L}_s = \| \mathbf{p}_{\text{leg}} - \mathbf{p}_{\text{leg}}^\text{ref} \|^2
\end{equation}
where \( \mathbf{p}_{\text{leg}} \) represents the actual swing leg positions and \( \mathbf{p}_{\text{leg}}^\text{ref} \) represents the reference swing leg positions derived from the quadratic polynomial fit.

\subsection{Coefficient Values}

The coefficients for each loss component in the locomotion policy loss function are set as follows:
$
\alpha_v = 5,
\alpha_R = 2,
\alpha_\omega = 1,
\alpha_h = 10^2,
\alpha_g = 1,
\alpha_q = 4,
\alpha_\tau = 10^{-3},
\alpha_s = 10^2.
$

\end{appendices}
\end{document}